# AutoPaint: A Self-Inpainting Method for Unsupervised Anomaly Detection

Mehdi Astaraki[1,2,4], Francesca De Benetti[3], Yousef Yeganeh[3], Iuliana Toma-Dasu[2,4], Örjan Smedby[1], Chunliang Wang[1], Nassir Navab[3,5], Thomas Wendler[3,6]

[1] KTH Royal Institute of Technology, Department of Biomedical Engineering and Health Systems, SE-14157 Huddinge, Sweden
[2] Karolinska Institutet, Department of Oncology-Pathology, SE-17176 Stockholm, Sweden
[3] Chair for Computer Aided Medical Procedures and Augmented Reality, Technische Universitat München, Boltzmannstr. 3, 85748 Garching bei München, Germany
[4] Stockholm University, Department of Physics, SE-106 91 Stockholm, Sweden
[5] Chair for Computer Aided Medical Procedures Laboratory for Computational Sensing and Robotics, Johns-Hopkins University, Baltimore, MD, USA
[6] SurgicEye GmbH, Munich, Germany

*Abstract*

Robust and accurate detection and segmentation of heterogenous tumors appearing in different anatomical organs with supervised methods require large-scale labeled datasets covering all possible types of diseases. Due to the unavailability of such rich datasets and the high cost of annotations, unsupervised anomaly detection (UAD) methods have been developed aiming to detect the pathologies as deviation from the normality by utilizing the unlabeled healthy image data. However, developed UAD models are often trained with an incomplete distribution of healthy anatomies and have difficulties in preserving anatomical constraints. This work intends to, first, propose a robust inpainting model to learn the details of healthy anatomies and reconstruct high-resolution images by preserving anatomical constraints. Second, we propose an autoinpainting pipeline to automatically detect tumors, replace their appearance with the learned healthy anatomies, and based on that segment the tumoral volumes in a purely unsupervised fashion. Three imaging datasets, including PET, CT, and PET-CT scans of lung tumors and head and neck tumors, are studied as benchmarks for evaluation. Experimental results demonstrate the significant superiority of the proposed method over a wide range of state-of-the-art UAD methods. Moreover, the unsupervised method we propose produces comparable results to a robust supervised segmentation method when applied to multimodal images.

1. Introduction

Medical image segmentation refers to the process of partitioning the voxels/pixels of tissues, organs, or pathologies from background anatomical structures in medical images such as Computed Tomography (CT) or Positron Emission Tomography (PET). Segmentation is recognized as one of the most challenging tasks in medical image analysis due to the complexity and variability of human anatomy, the lack of intensity/textural contrast between adjacent tissues, the variability of intensities in medical images, and the presence of noise/artifacts (Hesamian et al., 2019). As a result, this process is often done manually by clinical experts, which is not only demanding but also subject to inter/intra-observer variabilities (Fournel et al., 2021). However, the quantifications derived from the segmentation step deliver critical information regarding the characteristics of the segmented regions such as shape, area/volume, and intensity/textural distributions that can be further used for diagnosis, prognosis, and interventional purposes. In the context of oncological images, the aim of image segmentation is commonly to delineate the boundaries of target tumoral regions (Wadhwa et al., 2019) and/or nearby healthy organs (Fu et al., 2021).

In the past three decades, a variety of computerized methods have been developed to partially automate and speed up the delineation time without compromising the segmentation accuracy. In a broad view, these methods can be categorized as either deep learning (DL) or non-deep learning techniques. In the context of non-deep learning techniques, a wide range of rule-based methods have been proposed for different segmentation tasks. Region-growing (Thakur and Shyam Anand, 2004), watershed (Benson et al., 2015), level-set (Astaraki et al., 2018), Markov random fields (Goubalan et al., 2016), graph cut (Chen and Pan, 2018), atlas-based (Candemir et al., 2016), and statistical shape modeling (Chowdhury et al., 2012) approaches, are only a few examples of rule-based segmentation methods that were employed to segment different types of tumors in different organ systems such as liver (e.g., (Siriapisith et al., 2020; Zheng et al., 2018)), kidney (e.g., (Khalifa et al., 2017; Torres et al., 2018)), and prostate (e.g., (Delpon et al., 2016; Wong et al., 2016)).

*Supervised segmentation: Capability and limitations*. Thanks to the rapid advances in the DL fields, and in particular, convolutional neural networks (CNNs), a great level of progress has been witnessed in the performance of medical image segmentation tasks. Inspired by the breakthrough of the U-Net model (Ronneberger et al., 2015), many different DL-approaches have been proposed to tackle a variety of segmentation challenges. The novelties introduced by such models are mainly focused on modifications of the network architecture and/or optimization process. In this context, Attention U-Net was proposed by integrating the attention gate (Schlemper et al., 2019) into the plain U-Net model to guide the learning process more on the target area that successfully improved the segmentation performance of brain tumors (Islam et al., 2020) and retinal vessels (Zhang et al., 2019). By replacing convolutional blocks with inception blocks (Szegedy et al., 2015), computationally efficient deeper U-Nets were developed to deal with large variations in size and morphology within the salient regions. The superiority of the segmentation accuracy of such models was reported, e.g., for the challenging task of lung nodule detection (Cheng et al., 2019). Similarly, Dense U-Net and Residual U-Net were developed by using Dense blocks (Huang et al., 2016) and Residual blocks (He et al., 2016), respectively, in the encoder-decoder paths that lead to outstanding segmentation accuracy, e.g., of the prostate (Baldeon-Calisto and Lai-Yuen, 2020) and lung cancer (Azad et al., 2019). More powerful segmentation network families such as U-Net$^{++}$ (Zhou et al., 2018), Adversarial U-Net (Xue et al., 2018), and Swin-Unet (Cao et al., 2023) have been developed and tested on large-scale datasets with remarkable improvement in segmentation performance in different tasks. Finally, nnU-Net (Isensee et al., 2020) was developed as a self-configuring generic segmentation pipeline. Although the model architecture follows the conventional encoder-decoder structure, by carefully designing the preprocessing steps, hyper parameter tuning, and optimization process, they managed to outperform many existing models including highly specialized solutions on 23 different biomedical image segmentation challenges. Despite the promising potential of such models, which can achieve clinical expert-level accuracies, they require a large number of labeled data due to their supervised training fashion. In fact, supervised training of such data greedy models suffers from two types of limitations. First, the number of training medical images is often limited because of the costly slice-by-slice data annotation and the lack of large publicly available datasets. Second, even if large-scale training data is available, the generalization power of the learned models is limited to the class of data used for training. In fact, the variety of imaging protocols and infrastructure, as well as biological diversity necessarily requires the collection of annotated data for the particular problem to be tackled followed by retraining of the model to handle these cases of domain shift (Hansen et al., 2022).

*Unsupervised segmentation*. Unsupervised deep learning methods tend to be a preferable choice for medical image analysis tasks as their optimizations do not entail labeled datasets. In this domain, unsupervised anomaly detection (UAD) is an active field of research that aims to identify the data that does not fit the learned distribution from normal data (e.g., (Schlegl et al., 2019)). The main advantage of UAD approaches is their similarity to the learning procedures of physicians who are trained to learn the appearance and characteristics of healthy anatomical structures to potentially detect any arbitrary abnormalities without a-priori knowledge of their attributes (Baur et al., 2021). This essentially means that the training process of such models requires only unlabeled data acquired from healthy subjects. The underlying hypothesis is to capture the distribution of healthy anatomical organs by training deep representation learning models in order to identify anomalies as outliers with respect to the normative distribution (Baur et al., 2019). In the domain of medical image segmentation, the applications of UAD techniques have been extensively investigated for the task of lesion segmentation (e.g., (Baur et al., 2020a; Tian et al., 2021; Zimmerer et al., 2019)). In a series of contributions, Baur et al. investigated the potential of the deep autoencoder (AE) models for unsupervised brain lesion segmentation from magnetic resonance (MR) images (Baur et al., 2021). Specifically, by integrating the adversarial training into spatial variational AE (VAE), they could map the healthy anatomies into latent manifolds and further reconstruct fairly high-resolution images. This model was employed for multiple sclerosis (MS) lesion segmentation in a dataset containing 49 subjects (Baur et al., 2019). They later developed a SteGANomaly (Baur et al., 2020a) model, which gains from the steganographic abilities of CycleGAN in removing high-frequency patterns that, to some extent, was a beneficial strategy for preventing the learned model from reconstructing the images with pathological regions. The same authors employed the inherent multi-scale nature of the Laplacian pyramid within a family of AE models to compress and reconstruct MR images of different resolutions in a scale-space (Baur et al., 2020b) approach. Schlegel (Schlegl et al., 2019) built a generative model of healthy training data and used the GAN's latent space along with an anomaly score to comprise a discriminator feature residual error and image reconstruction error. The proposed f-AnoGAN model was tested on optical coherence tomography images showing superiority over conventional AE-based models. To efficiently learn fine-grained feature representations, Tian (Tian et al., 2021) developed a Constrained Contrastive Distribution (CCD) model to simultaneously predict the augmented data as well as image contexts. This model was tested on colonoscopy and fundus screening datasets and outperformed a few other UAD models. Naval (Naval Marimont and Tarroni, 2021) lifted the need for an encoder network to capture the latent representation of healthy data by substituting the AE architecture with an auto-decoder along with a modified version of the implicit field learning technique to reconstruct high-resolution anomaly-free images. This model was tested on a brain tumor segmentation task in MR images with an outstanding performance against a family of VAE models. Last but not least, Dey (Dey and Hong, 2021) developed an adversarial-based selective cutting neural network (ASC-net) by integrating the adversarial learning into a U-Net-like model with two decoders to decompose the images into two cuts based on a reference learned distribution of healthy images. The focus of this model is to obtain a joint estimation of anomaly and the corresponding normal images rather than reconstructing a high-fidelity normal-looking image. This model was tested on several different pathology segmentations, including MS and brain tumors in MR images and liver tumors in CT images, and outperformed the segmentation accuracy of the AnoGAN families.

*Anomaly detection challenges.* Despite the promising results achieved by the current UAD models, such models suffer from a number of limitations: 1) The first issue is related to learning the distribution of healthy anatomies in full image resolution. In fact, there are many fine-grained details in healthy anatomies that pose similar attributes with respect to the pathologies. However, the current methods cannot deal with such anatomical details and are unable to discriminate fine-grained healthy structures from abnormalities. To tackle this issue, the current methods normally reduce the dimensionality of the original images to eliminate the fine-grained details and train the models with low-resolution data (Baur et al., 2021). Such a downsampling procedure, however, abandons important image characteristics and therefore yields in learning the distributions of incomplete anatomies. 2) Another important limitation of the current UAD techniques is their difficulties in preserving the anatomical constraint within the generated images. In fact, generating a healthy image from the corresponding pathological image does not necessarily guarantee the retaining of the anatomical constraints of other tissues and structures. Therefore, the residual images calculated from the difference between the original images and the unrealistic-looking generated images often consist of many false positives. 3) Finally, conventional UAD models often focus on detecting anomalies that appeared with hypo or hyper-intensity patterns with respect to nearby normal tissues, such as glioma and MS lesions, in FLAIR MR sequences. However, to the best of our knowledge, detecting pathologies with similar intensity/textural patterns w.r.t. adjacent healthy organs has not been investigated thoroughly. The fact that AE models often reconstruct a blurry version of the down-sampled original image challenges the underlying hypothesis of capturing the distribution of healthy anatomies (Meissen et al., 2021). In other words, the hypo or hyper-intensity patterns of the studied pathologies within generated images from the learned low-dimensional representation space naturally tend to be suppressed. Hence, the residual images followed by some thresholding would consist of the anomaly regions regardless of how well the model could replace the pathologies with healthy tissues.

*Image inpainting.* Image inpainting is the process of synthesizing alternative contents in the missing parts of an image with semantically meaningful patterns to reconstruct a seamless and realistic-looking image. It can be used for a variety of image editing tasks such as text removal, object removal, and missing part recovery (e.g., (Elharrouss et al., 2019; Jam et al., 2021)). Although a variety of CNN-based models have been proposed for image inpainting, typical convolutional operators are naturally unsuitable for hole-filling as they treat all the valid and invalid pixels as the same. To tackle this issue, Liu (Liu et al., 2018) proposed a partial convolution (PConv) neural network in which the typical convolution operator is masked and renormalized to be conditioned only on the valid pixels. The invalid pixels are replaced by adjacent textures following a rule-based mask-updating procedure. The model was trained with randomly generated irregular masks, and its superior performance was verified on large-scale datasets both quantitatively and qualitatively. In order to condition the prediction of missing pixels at each coordinate on the valid pixels from the input image, Yu et al. (Yu et al., 2019) replaced the PConv layers with gated convolution (GConv) layers along with adding a contextual attention layer and spectral normalized markovian discriminator (SN-PatchGAN). The advantage of this GConv operator is that it can learn features from input images progressively for each channel of the network. The network architecture consists of two encoder-decoder networks named as coarse and refinement networks, followed by a fully convolutional SN-PatchGAN. Due to the learnable dynamic mask updating procedure, the GConv model generates images with more color and texture consistency than the PConv model. Last but not least, different studies show that inpainting models trained with irregular-shaped holes distributed randomly over the image plane can generate images with more semantic context than those trained with simple-shape holes such as rectangles (Liu et al., 2021; Wang et al., 2021).

*Contribution.* In this study, we propose an inpainting-based UAD method, AutoPaint, for tumor segmentation in single/multimodal medical images (The source code is available at https://github.com/XXX/XXX). Specifically, (1) we propose a robust inpainting method to reconstruct high-resolution medical images from corrupted ones while preserving fine-grained details. To efficiently train the inpainting model, healthy images were corrupted by generating random irregular holes to simulate the morphological characteristics of heterogeneous tumors. (2) The learned model is then employed for automatic tumor removal (and thus anomaly detection) in the test phase in an autoinpainting pipeline. In particular, a set of subregions within the main image is defined through a sliding window approach to be inpainted. (3) The autoinpainting procedure is followed by a postprocessing strategy to detect the candidate region for the final tumor removal. Finally, image slices of each subject are aggregated to form a volume from which residual volumes are calculated to segment the tumor volumes. The proposed inpainting model is optimized with a multi-term objective function to fill the invalid holes with plausible imagery characteristics as well as to preserve the anatomical constraints. The developed pipeline for unsupervised segmentation was tested with two types of tumors: Lung Cancer (LC) and Head-and-Neck (HN) cancer on single modalities of CT and PET as well as multimodal (two-channel) PET-CT images.

## 2. Methods

### 2.1. Dataset

Three datasets were examined to investigate the potential of the proposed method for segmenting different types of tumors.

#### 2.1.1. Internal PET-CT dataset for LC tumor segmentation

This internal dataset includes 33 subjects, all diagnosed with non-small cell LC in stage III except three subjects who were categorized as stage I, II, and IV. All subjects were scanned with a Biograph 40 PET scanner (Siemens Healthineers) to acquire a first $^{18}$F-fluoro-2-deoxy-d-glucose (FDG)-PET-CT scan before the beginning of radiation therapy and a second one after a few weeks of treatment. In this dataset, the voxel spacing in the CT images was fixed to $(0.976×0.976×3) mm^3$, and to $(4.072×4.072×3) mm^3$ for the corresponding PET images. A semi-automatic segmentation tool based on the level-set algorithm was utilized to generate the ground truth mask (Wang et al., 2014). In specific, initial contours were set around the tumors by an experienced user to instantiate the intensity-based contour evolution algorithm. The final contours were then visually examined and manually refined by an expert radiologist with more than 10 years of experience.

#### 2.1.2. AutoPET challenge dataset for LC tumor segmentation

The second LC dataset was obtained from the *automated lesion segmentation in whole-body FDG PET-CT* (AutoPET) challenge (Gatidis et al., 2022). The training set of the challenge data comprises 1,014 FDG-PET-CT scans of patients with histologically proven malignant melanoma, lymphoma, or LC. The whole-body 3D volumes, mainly, extend from the skull base to the mid-thigh level. From this cohort, 169 subjects containing LC tumors were selected for further analysis. The slice thickness and in-plane voxel size of the co-registered PET-CT volumes are 3mm and 2.036mm respectively. Lesion labels were manually annotated by two experienced radiologists through visual assessment of PET and CT information also considering other relevant clinical reports.

*2.1.3. HECKTOR challenge dataset for HN tumor segmentation*

For the HN tumor segmentation, data from *HEad and neCK TumOR (HECKTOR)* segmentation challenge 2022 were employed (Oreiller et al., 2022). The training set of this multi-institutional image data consists of 524 FDG-PET and low-dose non-contrast-enhanced CT images (acquired with combined PET-CT machines). All the patients were histologically diagnosed with HN cancer and underwent radiation treatment often combined with chemotherapy. The segmentation labels of the co-registered PET-CT volumes include gross target volume of primary tumors (GTVp) and gross target volume of lymph node involvements (GTVn). The contours were manually delineated by an expert radiologist and cross-checked by another independent expert. In specific, the edges of the morphological anomalies visible on CT images along with the corresponding hypermetabolic volumes from the fused PET-CT visualizations were used to delineate the contours. In this dataset, the in-plane voxel size ranged from 0.488mm to 2.733mm and slice thickness varied from 1mm to 5mm.

## 2.2. Image preparation and preprocessing

The following preprocessing was applied to the employed datasets. First, on the internal LC dataset, a third-order Spline interpolation method was used for the PET images to resample the voxel spacing of PET data into the corresponding CT volumes. Second, the intensity values of PET images were converted into standardized uptake values (SUV). Third, to enhance the contrast between the tissues within the target organ, intensity values of CT and PET images were clamped. Particularly, the Hounsfield values of CT images were clamped into the range of [-1000,500] for LC data and [-200,200] for the HN dataset. The SUV values of PET images were constrained in the range of [0,12] as well. The axial slices from signed 16bit volumes were extracted and saved with the size of 512×512 pixels. For training the models, the intensity range of images was normalized by maximum values and rescaled into the range of 0 to 1. Finally, the original two-classes segmentation task of the HN tumors was converted into a binary segmentation task by considering both GTVp and GTVn as the single target. Figure 1 shows the diversity of shape, size, and location of the tumors among the employed datasets.

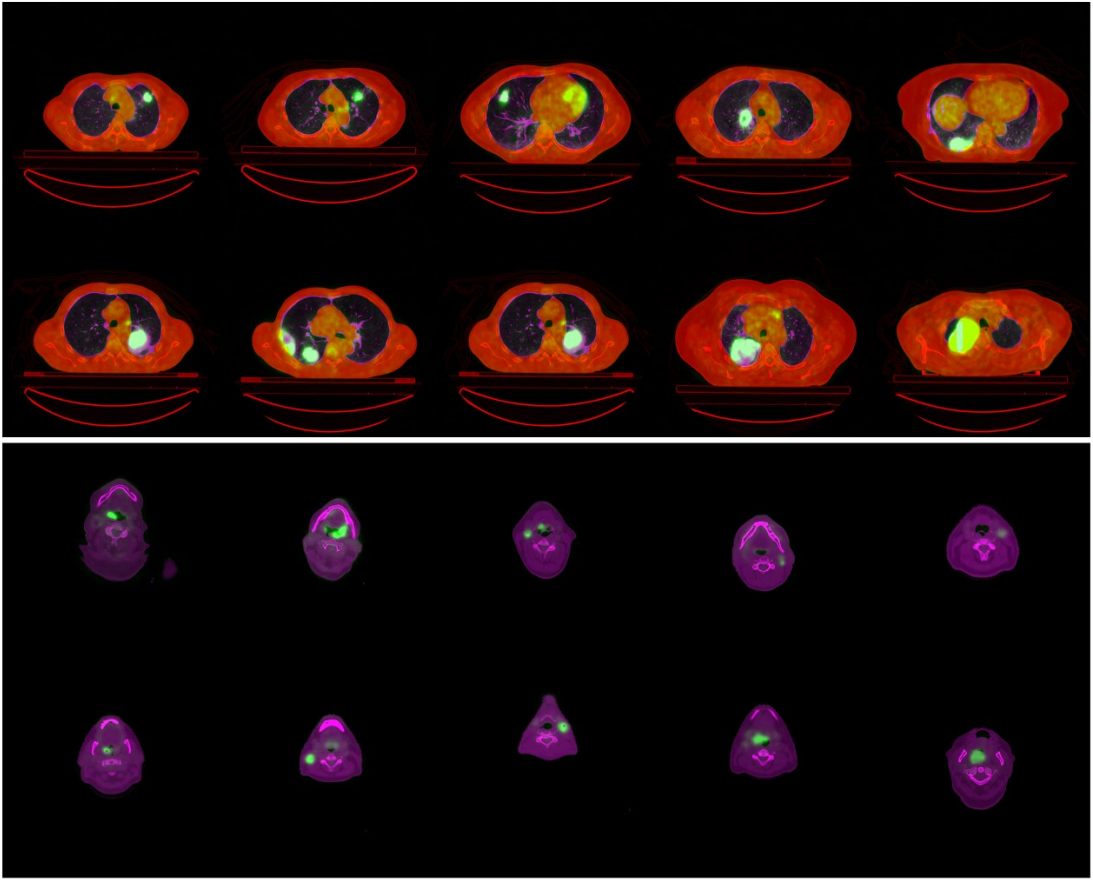

Figure 1. Heterogeneous tumors appear in a diverse range of shapes and sizes at different locations. The first two rows show the diversity of LC tumors, and the second two rows depict different HN tumors. In the lung dataset we use for the CT a red color scale, while for the PET a green one. For the HN dataset, we use magenta for the CT and again green for the PET.

### 2.3. Image inpainting model

Assume that $I_{(m_I, n_I)}$ and $W_{(m_w, n_w)}$ stand for a $c$-channel input image (or input feature map), and a filter, respectively. The conventional convolutional operator filters the input image and returns a $c'$-channel output, $O$. Mathematically, this function can be represented as:

$$O(i,j) = I_{(m_I, n_I)} * W_{(m_w, n_w)} = \sum_{m=0}^{(m_I-1)} \sum_{n=0}^{(n_I-1)} I(m,n).W(i-m, j-n)$$

where $0 \leq i < m_I + m_w - 1$, and $0 \leq j < n_I + n_w - 1$. Please note that for the simplicity of the notation, the bias term was skipped. Although this type of convolutional operator works well for several tasks such as image classification, segmentation, and detection, it is not suitable for the task of image inpainting. In fact, the sliding kernel scans all the pixels and elements within the image/feature maps and applies the same filters at different spatial coordinates. Thus, it simply ignores the presence of holes within a subregion and considers the valid and invalid pixels as the same. As a result, the inpainted holes do not fully match with the nearby textures, and the generated images contain textural/color inconsistencies.

The PConv operator (Liu et al., 2018) was proposed as a promising attempt to tackle the mentioned issues faced by the convolutional operators. Let $M$ be a binary mask with the same size as the input image, the partial convolution at every spatial coordinate for the current sliding window can be defined as:

$$O_{(i,j)} = \begin{cases} W^T_{(m_w,n_w)}(I_{(m_I,n_I)} \cdot M_{(m_I,n_I)}) \dfrac{sum(S_{(m_I,n_I)})}{sum(M_{(m_I,n_I)})}, & sum(M) > 0 \\ 0, & sum(M) \leq 0 \end{cases}$$

where $S$ is an all-one matrix with the same size of $M$. Compared to the ordinary convolution operator, one can understand that the output values of PConv depend only on the valid areas defined by the binary mask ($M$). Accordingly, if there exists even one single valid pixel within the subregion covered by the sliding kernel, the convolution operator will function. In this case, the central element of the corresponding subregion of the binary mask (M) will be updated as well. The role of the scaling factor ($\frac{sum(S)}{sum(M)}$) is to adjust for varying sizes of the valid regions. The rule-based procedure for updating the binary mask is problematic because: 1) all feature channels in each convolutional layer share the same mask regardless of their inconsistencies which is not optimal, especially for multi-channel input images such as multimodal PET-CT slices. 2) The binary mask will be updated progressively as it goes deeper into the network so that all the invalid pixels will disappear no matter how many pixels were covered in the previous layers.

The GConv operator (Yu et al., 2019) has been proposed to turn the problematic rule-based mask updating of PConv into a learnable procedure. In specific, gated convolutions learn soft mask updating automatically from the image/feature maps. It will enable the convolutional operators to learn the dynamic feature selection mechanism for each channel and each spatial coordinate independently. This process can be formulated as:

$$Gating_{(i,j)} = \sum\sum W_g \cdot I$$

$$Feature_{(i,j)} = \sum\sum W_f \cdot I$$

$$O_{(i,j)} = \varphi(Feature_{(i,j)}) \odot \sigma(Gating_{(i,j)})$$

where $\sigma$ refers to the sigmoid function that scales the output of the gating signal into the range of 0 to 1; $\varphi$ can be any kind of nonlinear activation function; $W_g$ and $W_f$ are two separate convolutional filters.

Inspired by the concept of the PConv model and GConv operator, in this study, we design a U-Net-like architecture, replacing all the ordinary convolutional layers with the GConv layer and using the nearest neighbor upsampling method in the decoder path. Specifically, the encoder part of the model consists of 8 GConv blocks, each of which includes a GConv layer with a stride of 2, followed by an optional batch normalization (BN) layer and a rectified linear unit (ReLU) activation function. The decoder stage of the model, similarly, contains 8 GConv blocks, each of which consists of a nearest neighbor upsampling layer, a GConv layer, an optional BN layer, followed by a Leaky ReLU activation function. The skip connections concatenate the feature maps and corresponding binary masks from the encoder blocks to the corresponding decoder blocks with the same resolution. The final output layer of the model is an ordinary convolutional layer with a sigmoid activation function which is fed by a concatenation of the last GConv block from the decoder path and the original input image with holes along with the original binary mask from the encoder. This strategy enables the model to directly transfer and copy the values of the valid pixels to the output layer. Figure 2 demonstrates a graphical illustration of the network architecture.

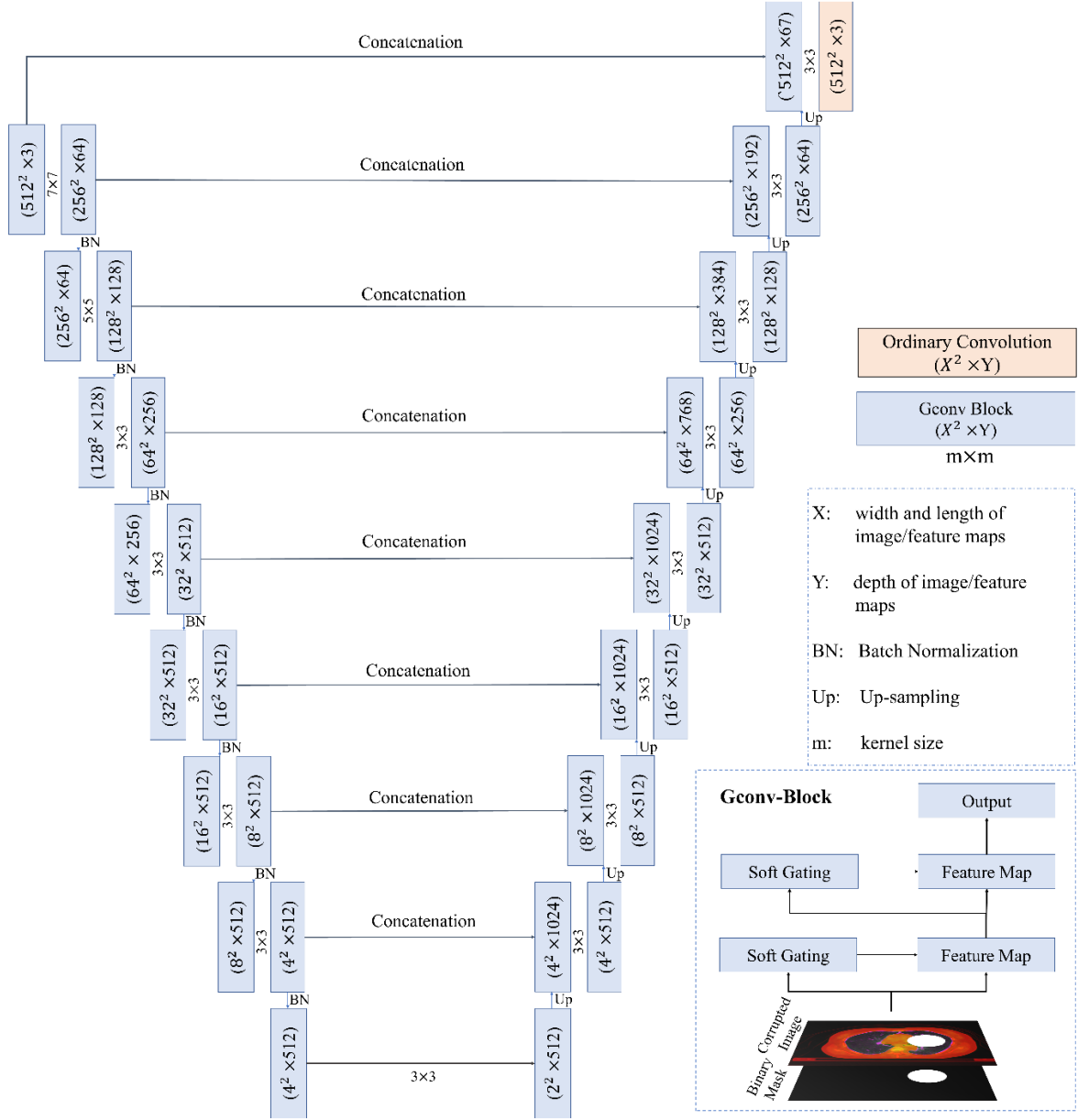

Figure 2. Schematic illustration of the model architecture.

In order to fill the holes with meaningful semantic patterns, the proposed model is optimized with a multi-term objective function (Liu et al., 2018) that takes into account both pixel-wise reconstruction accuracy and context information. Let the input image with holes be $I_{in}$; $I_{gt}$ represents the original image without holes (ground truth), $I_{out}$ indicates the predicted image, and $M$ denotes the binary mask used for corrupting the image; the first two terms in the objective functions are pixel-wise errors that can be calculated separately for the valid and invalid regions as the mean absolute errors ($L^1$ norm). These two terms aim to minimize the intensity differences between the predicted and ground truth images inside and outside the hole regions separately:

$$L_{valid} = \frac{1}{N_{I_{gt}}} \|M \odot (I_{out} - I_{gt})\|_1$$

$$L_{hole} = \frac{1}{N_{I_{gt}}} \|(1-M) \odot (I_{out} - I_{gt})\|_1$$

where $N_{I_{gt}}$ shows the number of pixels in the $I_{gt}$.

The third term is perceptual loss which aims to minimize the discrepancies between the high-level feature representations extracted from the predicted and ground truth images in order to maximize the perceptual similarity between these two images. It calculates the $L^1$ norm between two sets of high-level features extracted from $I_{out}$ and $I_{comp}$ where $I_{comp}$ is the composite output which is similar to the predicted image but with the intensity of valid pixels replaced by those of the ground truth. 1st, 2nd, and 3rd pooling layers of a pre-trained VGG16 (Simonyan and Zisserman, 2015) network were used to extract the features:

$$L_{perceptual} = \sum_{p=0}^{p-1} \frac{\|\Psi_p^{I_{out}} - \Psi_p^{I_{gt}}\|_1}{N_{\Psi_p^{I_{gt}}}} + \sum_{p=0}^{p-1} \frac{\|\Psi_p^{I_{comp}} - \Psi_p^{I_{gt}}\|_1}{N_{\Psi_p^{I_{gt}}}}$$

here, $\Psi_p^{I_*}$ refers to the outputs of the activation function of the *p*th layer of the pre-trained network given the input $I_*$.

To minimize the style differences between the synthesized and ground truth images, style loss was computed as well. To reconstruct images with high level of style similarities inside and outside of the holes, the style error was calculated for predicted and composite images separately:

$$L_{style_{out}} = \sum_{p=0}^{p-1} \frac{1}{C_p^2} \| \frac{1}{K_p} ((\Psi_p^{I_{out}})^T (\Psi_p^{I_{out}}) - (\Psi_p^{I_{gt}})^T (\Psi_p^{I_{gt}})) \|_1$$

$$L_{style_{comp}} = \sum_{p=0}^{p-1} \frac{1}{C_p^2} \| \frac{1}{K_p} ((\Psi_p^{I_{comp}})^T (\Psi_p^{I_{comp}}) - (\Psi_p^{I_{gt}})^T (\Psi_p^{I_{gt}})) \|_1$$

The style loss is similar to the perceptual loss, but it first calculates the autocorrelation of extracted features and then computes the $L^1$ norm. In this notation, $C_p$ indicates the depth of the channels in $\Psi_p$, and $K_p$ refers to the number of elements in $\Psi_p$ tensor.

The sixth loss term is total variation (TV) which is a conventional objective function for noise reduction applications. In fact, it functions as a smoothing term that makes the intensity values of the neighboring pixels in the synthesized image closer to each other:

$$L_{tv} = \sum_{(i,j) \in R, (i,j+1) \in R} \frac{\|I_{comp}^{i,j+1} - I_{comp}^{i,j}\|_1}{N_{I_{comp}}} + \sum_{(i,j) \in R, (i+1,j) \in R} \frac{\|I_{comp}^{i+1,j} - I_{comp}^{i,j}\|_1}{N_{I_{comp}}}$$

where $N_{I_{comp}}$ is the number of pixels in the composite image.

Finally, since the early layers of the model focus on capturing edge-based features, the described pixel-wise, perceptual, style, and TV losses alone cannot well preserve the high-frequency patterns. This issue will be problematic when the contents of each channel of the input image carry different structures, such as multimodal PET-CT images. Accordingly, to maintain the edges and synthesize images with details as much as possible, the last term includes the Laplacian (lap) pyramid loss:

$$L_{lap(I_{out}, I_{gt})} = \sum_j 2^{2j} \|L^j(I_{out}) - L^j(I_{gt})\|_1$$

where $L^j(x)$ refers to the *j*th level of the Laplacian pyramid representation of input *x*. In this study, the parameter *j* was set to 3, i.e., three levels of pyramid representations were computed.

Therefore, the overall objective function is the combination of all the mentioned loss terms:

$$L_{total} = 30L_{valid} + 240L_{hole} + 0.2L_{perceptual} + 0.05(L_{style_{out}} + L_{style_{comp}}) + 250L_{tv} + 20L_{lap}$$

The coefficient of each term was fixed after conducting an ablation study over 2000 test images (see section 2.1 in Supplementary Materials).

### 2.4. Learning the appearance of normal anatomies

The proposed inpainting model was employed to learn the attributes of healthy anatomical structures by learning to fill the irregular holes with the characteristics of healthy structures. In other words, healthy image slices corrupted with irregular random holes are used to train the inpainting model. Having the corrupted healthy images as input to the model on one side and the original healthy images as the ground truth on the other side, the inpainting model is trained to smoothly replace the holes with semantically meaningful patterns in order to synthesize realistic-looking images while preserving fine-grained details and anatomical constraints. With this strategy, the inpainting model is assumed to estimate the distribution of healthy anatomies.

Considering the fact that tumors appear with irregular shapes and different sizes at different locations, the corrupting holes should be generated in a way to imitate the visual attributes of the tumors. Accordingly, irregular holes were synthesized by carefully combining ordinary regular geometric shapes, including circles, ellipses, and lines. Thus, the simulated holes were distributed randomly over different spatial coordinates of the image space to occupy, on average per batch, 25 to 30 percent of the image size. With this approach, two models were trained separately for LC and HN datasets. In specific, 9000 healthy images from the AutoPET LC dataset and 15500 healthy slices from the HECKTOR HN dataset were extracted to train the inpainting model. An additional 2000 slices from each dataset were used as the validation set. To avoid data leakage, slices of patients used in the training set were not used in the validation set.

Each model was trained for 300 epochs with an Adam optimizer and a batch size of 8. The presence of the holes in the image causes issues with the BN parameters updating because the zero values inside the holes will contribute to updating the mean and variance of BN. Accordingly, it sounds rational to disable the calculation of the BN inside the holes. On the other hand, the training procedure forces the model to gradually fill the holes until they completely disappear so that they can potentially contribute to the BN parameter updating. Hence, the training was done in two phases. In the first phase, the models were trained for 150 epochs with a learning rate of 0.0001 and enabled all the BN layers. In the second phase, the model continues training for another 150 epochs with a learning rate of 0.00005. In this phase, the BN layers within the encoder path were disabled while they were active for the decoder stage. This fine-tuning strategy is not only beneficial to speed up the convergence but also to avoid the incorrect calculations of the mean and variance parameters of the BN operator (Gruber, 2019; Liu et al., 2018). The accuracy metrics over the validation set were monitored, and a certain epoch that resulted in the best accuracy metrics was used for the testing phase. It is worth mentioning that the described training procedure was performed independently for each of the examined imaging modalities, i.e., CT, PET, and PET-CT scans. Figure 3 demonstrates the qualitative performance of the model in replacing the irregular holes with the appearance of normal anatomical regions.

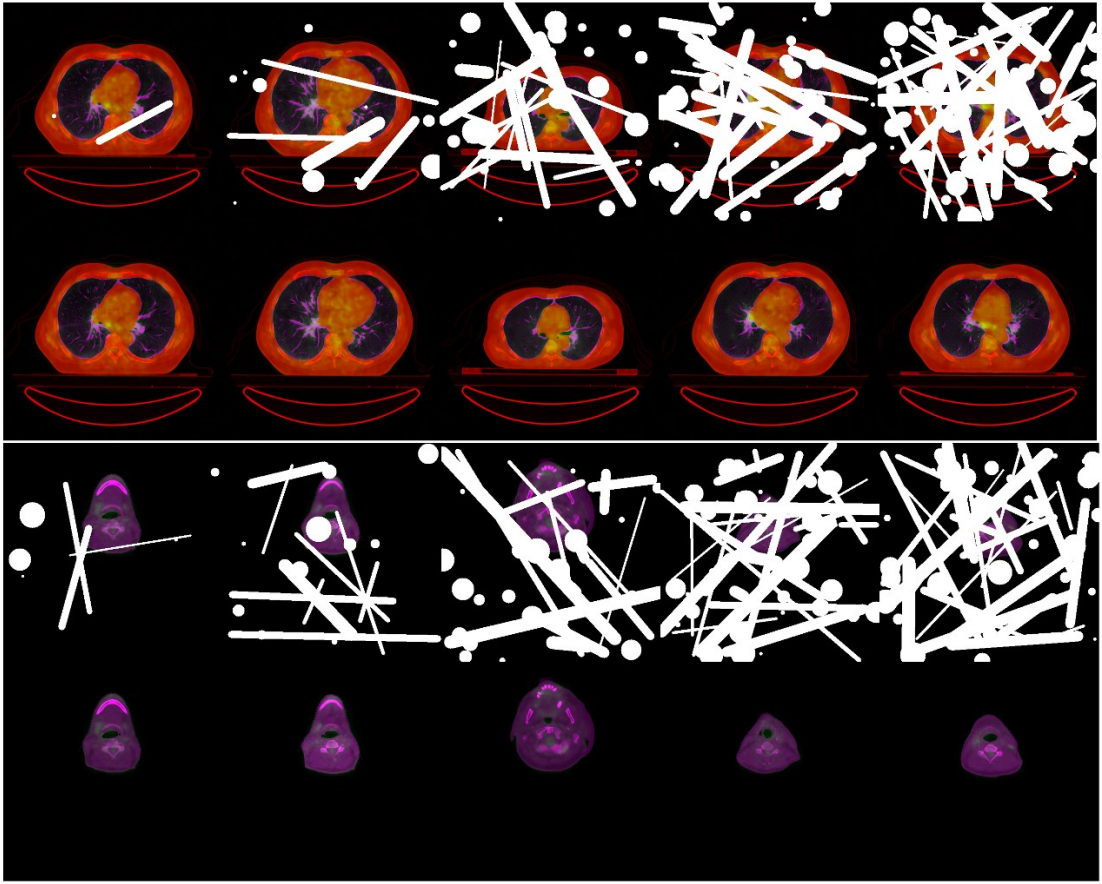

Figure 3. Examples showing how the inpainting model could successfully replace the irregular random holes with the appearance of healthy anatomies while preserving the anatomical constraints. For each set of LC and HN tumors, the first row shows the corrupted images with random holes (test images), and the second row illustrates the inpainted results in the inference phase.

## 2.5. Autoinpainting for unsupervised tumor segmentation

The trained inpainting model learns to synthesize semantically correct and contextually smooth contents in the predefined missing regions. Training the model only with healthy slices reinforces the model to replace the missing healthy tissues with the appearance of healthy tissues. This strategy enables the inpainting network to model the distribution of healthy anatomical structures that can be further utilized to detect anomalies as outliers from the learned distribution. In other words, replacing the tumor with the appearance of already learned healthy tissues leads to synthesizing tumor-free images from which the tumoral regions can be detected by calculating the differences between the original and synthesized images. Accordingly, the learned inpainting network, which was trained only with random holes, can function as a UAD model, given that no segmentation label is required to localize the tumor location. That being the case, a pipeline is proposed to turn the manual inpainting network into an autoinpainting model to segment the tumors in an unsupervised fashion.

The underlying idea thereby is to replace the random holes with a sliding window to sweep different anatomical regions for the inpainting process. Therefore, if the sliding window covers healthy regions, the inpainting network will replace the appearance of healthy structures with learned healthy structures; thus, the newly generated images remain intact. On the other hand, if the sliding window encounters tumoral regions, it substitutes the textures of the tumors with the appearance of already learned healthy tissues. Accordingly, for each original tumoral slice, a fake tumor-free image can be generated without needing any kind of supervised signal. Hence, a pipeline is proposed to efficiently inpaint the tumoral regions while preserving the appearance of healthy tissues with anatomical constraints. This pipeline consists of the following four steps: I) preparing the input slices, II) detecting the candidate regions, III) determining the target region, IV) segmenting the target tumor:

*Preparing the input slices*

The employed AutoPET and HECKTOR datasets contain other anatomical organs in addition to the target chest and neck regions. Therefore, to concentrate the analyses within the target organs, extended lung field masks for LC datasets and HN masks for HN data were delineated. In specific, a pretrained progressive holistically-nested networks (P-HNNs) (Harrison et al., 2017) was used for the CT volumes to segment the lung fields in the presence of pathologies. The segmentation masks were visually examined, and manual refinements were needed only for a very limited number of cases. The masks were then dilated by morphological operators to roughly estimate the lung locations within the volumes. Finally, the dilated binary masks were applied to the corresponding PET images as well. For the HECKTOR dataset, the oropharyngeal regions were automatically detected by a model proposed by Andrearczyk (Andrearczyk et al., 2020). The bounding boxes were manually examined and refined for those subjects with out-of-boundary heads, tilted heads, or low SUVs. These preprocessing steps assure us that all further analyses will be performed within the Organ Of Interest (OOI) where the potential tumors are presented. The final preparation step includes the extraction of all the axial slices from the OOIs.

*Detecting the candidate regions*

Depending on the size of the OOIs, a certain number of subregions is determined with the help of a sliding window strategy for further analyses. In particular, a sliding circle sweeps over the OOIs in each of the axial slices. The sliding circle has a radius of 27 pixels and an interval distance of 15 pixels for LC, a radius of 15, and an interval distance of 8 pixels for HN datasets. In our experiments, we noticed that masks with circular shapes can perfectly cover the tumoral regions. In addition, circular shapes were already used in the training phase to create random holes. Therefore, the shape of the sliding windows was set as the circles. The already trained network is employed as an inference model to inpaint each of the moving circles independently. In other words, the sliding window scans each slice to produce several candidate circles to be inpainted by the trained network. The inpainting model, therefore, replaces the contents of the coordinates occupied by the circles with the textural patterns it learns from the healthy images in the training phase. As a result, for each of the circles within one slice, there will be a new synthesized image. If the moving circular masks a healthy subregion, the inpainting model replaces it with the texture of healthy tissues, and therefore there will be no remarkable intensity and textural differences between the original and the synthesized images. On the other hand, if the moving circle masks a tumoral region, the learned inpainting model replaces the textures of the tumor with the patterns of healthy tissues. Such a replacement leads to observing notable intensity and textural differences between the input slice and the generated slice. Accordingly, to identify which of the moving windows could cover the anomalies, the intensity and textural differences between the input slices and the inpainted slices were calculated. Specifically, for each of the moving circles, the differences between the original images and the inpainted images in the intensity domain (intensity difference), and feature map domain (textural difference) were calculated. These values were then sorted, and only the top few values with notable differences w.r.t. the other values were kept as they represent notable changes between the input image and the synthesized one, which could potentially imply the anomaly location.

It should be emphasized that if the size of the moving window is too small, the inpainting model will not be able to completely replace the tumoral regions. On the other hand, if this size is too large, it may slightly change too many tiny details, which would slightly change the general context. Thus, this size should be defined as a trade-off between the largest and smallest possible tumors within the datasets. In this study, based on the diversity of tumor sizes, a range of potential values were examined in an ablation study which yields setting the radius of the moving window equal to 27 and 15 for the two studied tumors as the optimal value (See section 2.2 in Supplementary Materials).

*Determining the target region*

The identified top candidate regions either masked one single tumor or covered different anomalies related to multi-focal tumors. To automatically find out whether the top candidate regions share the same tumor or they focus on various subregions, the union of the top candidate binary masks is calculated. To this extent, if the top candidate regions overlap each other, their union will form a larger binary mask; however, if they do not share even a single pixel, the outcome of the union calculation will not differ from the originally separated masks. This scheme estimates whether only one tumor or several tumors are presented in the slice. Then, the updated union mask will be ready to perform the final inpainting step. Considering the possibility of the presence of extremely large-size tumors, this final mask may not be large enough to cover the whole abnormalities. Accordingly, the size of this binary mask needs to be enlarged without compromising the efficacy of small-size anomalies. To do so, an incremental morphological dilation approach is adopted in order to dilate the updated binary mask with structural elements of the width of [7,9,11,13,15]. Simply explained, in addition to the updated union mask, five other dilated versions of this mask will be generated to conduct a total number of six final inpaintings independently. For each of them, the textural and intensity differences between the input slice and the inpainted slices will be quantified, and if no changes are observed between the sequential orders, then the mask with the smaller size will be selected; otherwise, the one with the larger size will be set as the final candidate mask(s). In this way, the small-size tumors will not be affected by this incremental dilation strategy as they remain inpainted with the updated union mask, while the extremely large-size tumors can be covered more efficiently by the dilated masks.

*Segmenting the target tumor*

The proposed pipeline analyzes all the axial slices in the OOI; however, not all the slices contain tumors. Therefore, to prevent the model from detecting small deviations in healthy slices as anomalies, a size-based criterion is integrated into the pipeline. In fact, the radius of the smallest tumor in the studied dataset was measured to be 7 pixels for LC and 4 pixels for HN tumors. Having known the minimum values, any detected abnormalities with sizes smaller than the minimum radius can be recognized as a false positive and skipped from the further steps. To implement this concept, first, the residual images are calculated as the differences between the input and the final inpainted images. Connected components (CCs) of the residual images are computed, and the size of the largest CC at each slice is compared against the minimum radius of the tumors. If the condition is satisfied, the output of the algorithm will become the final inpainted slice; otherwise, the input slice will be directly set as the output. The latter case necessarily means that either the model could not detect the tumor(s) or the image slice does not contain any tumors. Figure 4 illustrates a general schematic presentation of the autoinpainting pipeline. Please note that the segmentation of LC tumors in CT images is more challenging than the PET modality or PET-CT multimodal images; therefore, the graphic illustration in figure 4 is depicted on a CT slice to accentuate the abilities of the proposed pipeline.

The mentioned process is repeated for all the axial slices from which a stack of volume can be formed from the algorithm outputs. Therefore, for each input volume, there will be a synthesized autoinpainted volume. The intensity range of both input and synthesized one lies in the range of 0 to 1. The final residual volume is then computed as intensity differences between the two volumes. Finally, to quantify the segmentation performance, two approaches were followed. First, a variable thresholding value in the range of 0 to 0.8 with an incremental rate of 0.02 was used to binarize the residuals for further quantifications, from which the threshold that leads to the best segmentation accuracy for that subject was selected. Therefore, a subject-specific threshold value is used for the quantification. These metrics are reported by the ∏ notation. Second, a conventional quantification was done by setting a single threshold value to binarize all the residual volumes. In particular, from the variable threshold range, the one that yields the highest segmentation accuracy over all subjects was chosen as the fixed thresholding value. This fixed value can be used for the inference phase.

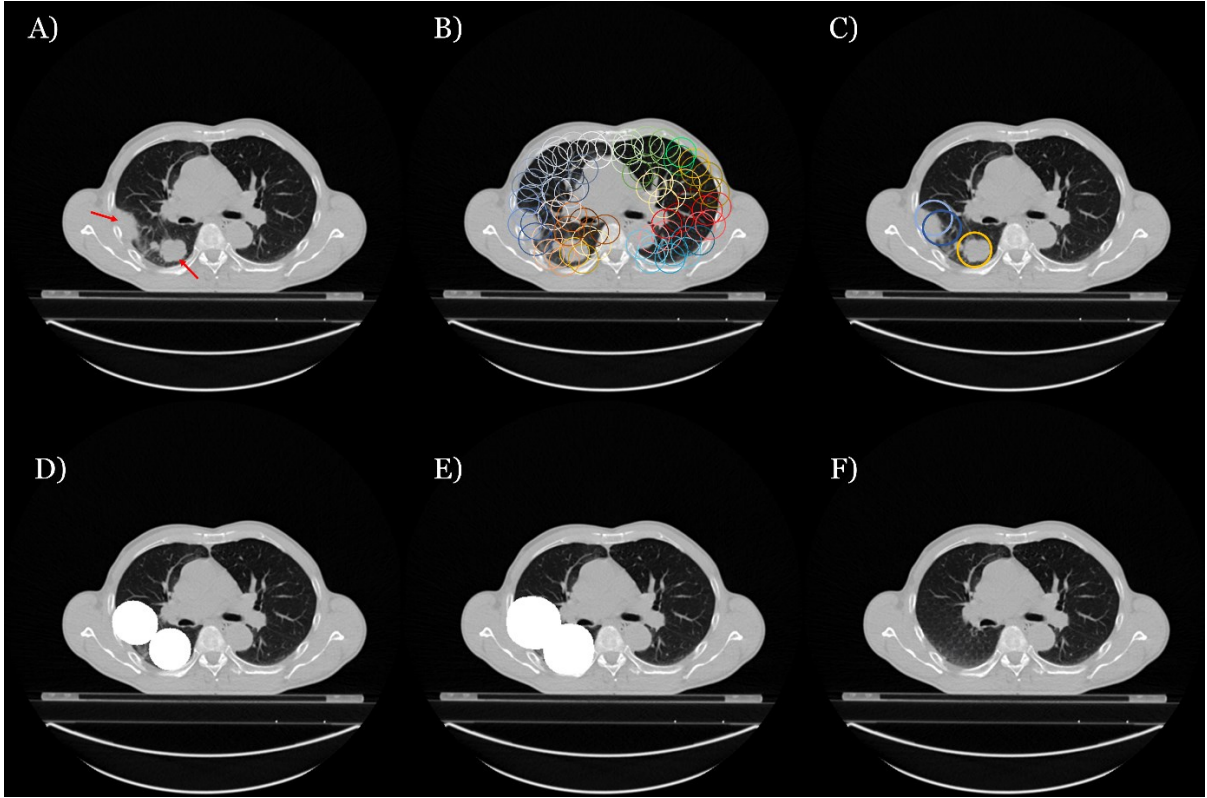

Figure 4. The autoinpainting pipeline employs the moving window strategy to adaptively inpaint the tumoral regions in a pure unsupervised approach. A) the original slice consists of multi-focal tumors (depicted with red arrows); B) the determined circles to be inpainted independently are presented with different colors; C) the top three candidate circles detected the two different tumors; D) the union of candidate regions was used to corrupt the image for inpainting process; E) incrementally increasing the size of the detected regions better cover the tumoral zones and F) final inpainted image does not contain the tumors anymore.

## 2.6. External validation

To benchmark the efficacy of the proposed method, we compare its performance to state-of-the-art (SOTA) models in two folds: 1) A supervised segmentation model was employed to find out what optimal performance can be achieved over the investigated tasks. In specific, the self-configuring nnU-Net model (Isensee et al., 2020) as a powerful segmentation framework was utilized to estimate the optimal achievable segmentation accuracy of the studied dataset. This model was trained with a 5-fold cross-validation fashion for each dataset separately. The default settings of the nnU-Net framework were adopted without further modifications, and the models were trained for 1000 epochs. 2) A set of recently developed deep UAD models were analyzed as well to objectively compare the segmentation accuracy of the proposed unsupervised model against the relevant UAD references. In this context, the following models were examined (Baur et al., 2021): dense AE (dAE), spatial AE (sAE), context-encoding AE (ceAE), variational AE (VAE), context-encoding variational AE (ceVAE), Gaussian mixture variational AE (GMVAE), fast-AnomalyGAN (F-AnoGAN), and adversarial AE (AAE). For each of the datasets, healthy slices were used to train these UAD models, and the pathological slices were employed in the test phase. No slices of patients present in the training sets were used in the validation set.

## 2.7. Quantitative evaluation

To assess the performance of the proposed inpainting model and the segmentation pipelines, two sets of quantitative metrics were examined.

The first group of metrics includes mean square error (MSE), peak signal-to-noise ratio (PSNR), and structural similarity index (SSIM). These metrics are measured to quantitatively evaluate the performance of the inpainting network by directly comparing the original image to the synthesized one. The MSE metric measures the amount of changes per pixel between the two images; therefore, the smaller value of this measure represents more similarity between the two images. PSNR is another quality assessment measure between the two images where the higher PSNR value indicates the better quality of the synthesized image. SSIM assesses the perceptual image quality to quantify the visible differences between the two images. Let the original image be $I_{org}$, and $I_{out}$ shows the synthesized image with equal matrix sizes of m×n and the maximum possible intensity value of R; then, the metrics can be mathematically defined as:

$$MSE(I_{org}, I_{out}) = \frac{1}{m \times n} \sum_{i=0}^{m-1} \sum_{j=0}^{n-1} |I_{org}(i,j) - I_{out}(i,j)|^2$$

$$PSNR(I_{org}, I_{out}) = 10 \log_{10} \left( \frac{R^2}{MSE(I_{org}, I_{out})} \right)$$

$$SSIM(I_{org}, I_{out}) = \frac{(2\mu_{I_{org}} \mu_{I_{out}} + c_1)(2\sigma_{I_{org}I_{out}} + c_2)}{(\mu_{I_{org}}^2 + \mu_{I_{out}}^2 + c_1)(\sigma_{I_{org}}^2 + \sigma_{I_{out}}^2 + c_2)}$$

where $\mu_{I_{org}}$, and $\mu_{I_{out}}$ are average intensities; $\sigma_{I_{org}}^2$ and $\sigma_{I_{out}}^2$ are variance values and $\sigma_{I_{org}I_{out}}$ represents the covariance of the two images. Parameters $c_1$ and $c_2$ are two variables that ensure stability when the denominator becomes 0.

The second group of metrics is used to quantify the segmentation accuracy of the proposed pipeline. These metrics include the Dice coefficient (DSC), Precision, and Recall. While DSC measures the overlap between the target masks and model predictions, Precision and Recall metrics demonstrate the accuracy of pixel classifications. Given that $S$ represents the segmentation output of the model and $G$ refers to the ground truth mask, $T_p$, $F_p$, $F_N$ show true positive, false positive, and false negative, respectively, calculated from the confusion matrix. the definitions of the metrics are formulated as follows:

$$DSC = \frac{2|S \cap G|}{|S| + |G|}$$

$$Recall = \frac{T_p}{T_p + F_N}$$

$$Precision = \frac{T_p}{T_p + F_p}$$

## 3. Results

In this section, the performance of the proposed autoinpainting method for unsupervised tumor segmentation is presented in two folds: (1) the quality of the inpainting model, and (2) the segmentation accuracy of the autoinpainting pipeline.

### 3.1. Inpainting quality

There exist many possible solutions to quantify the performance of inpainting models; therefore, we employed the described MSE, PSNR, and SSIM metrics as conventionally have been used by other studies (Liu et al., 2018; Yu et al., 2019). Furthermore, qualitative comparisons are included by demonstrating both the corrupted and inpainted images. In the following, GConv$_{Lap}$ denotes the proposed method, which is compared against PConv and ordinary GConv models.

Tables 1, and 2 represent the comparison results between the performance of the models for each of the PET-CT, CT, and PET images for the LC and HN datasets separately. For the LC datasets, the trained model with AutoPET dataset was tested on the internal LC images for the metric quantifications. For the HN dataset, the quantified values come from the 2000 images from the validation set. In detail, the models trained with the healthy slices were used, in the test phase, to inpaint the corrupted validation/test images. Original images were then compared against the model predictions using the three quantitative metrics.

Table 1 – Numerical comparison between the performance of inpainting models on the internal LC dataset. The best-quantified metric for each of the model-data is marked in bold.

| Model-Data | Quantitative Metrics ($\mu\pm\sigma$) | | |
|---|---|---|---|
| | MSE↓ | PSNR↑ | SSIM↑ |
| PConv-CT | 123.401±66.536 | 27.915±2.623 | 0.908±0.033 |
| GConv-CT | 67.098±48.486 | 31.311±4.022 | 0.939±0.031 |
| GConv$_{Lap}$-CT | **66.041±47.330** | **31.495±4.332** | **0.943±0.030** |
| PConv-PET | 22.722±22.925 | 35.981±3.413 | 0.961±0.014 |
| GConv-PET | 21.931±28.111 | 37.449±5.094 | 0.973±0.015 |
| GConv$_{Lap}$-PET | **21.888±31.336** | **38.070±5.836** | **0.977±0.013** |
| PConv-Multi | 69.428±37.546 | 30.385±2.530 | 0.947±0.019 |
| GConv-Multi | 45.850±32.813 | 32.814±3.682 | 0.960±0.018 |
| GConv$_{Lap}$-Multi | **44.290±33.785** | **33.271±4.267** | **0.966±0.018** |

From Table 1, we can infer that the proposed GConv$_{Lap}$ model could inpaint the corrupted images more accurately than the other two models. In particular, the numerical metrics obtained from the proposed GConv$_{Lap}$ indicate lower error in terms of the MSE metric and higher similarity in terms of PSNR and SSIM for all the experiments regardless of the type of input images. As expected, quantitative values of the PET image show higher accuracy compared to those of the CT and multimodal images for all the experiments.

Table 2 – Numerical comparison between the performance of inpainting models on the HN dataset. The best-quantified metric for each of the model-data is marked in bold.

| Model-Data | Quantitative Metrics ($\mu\pm\sigma$) | | |
|---|---|---|---|
| | MSE↓ | PSNR↑ | SSIM↑ |
| PConv-CT | 9.934±8.367 | 39.922±4.561 | 0.985±0.012 |
| GConv-CT | 7.136±7.504 | 42.295±5.868 | 0.988±0.011 |
| GConv$_{Lap}$-CT | **5.744±6.177** | **43.622±6.396** | **0.991±0.009** |
| PConv-PET | 5.370±10.208 | 45.476±6.732 | 0.992±0.006 |
| GConv-PET | 4.270±8.621 | 46.462±6.660 | 0.991±0.007 |
| GConv$_{Lap}$-PET | **3.130±6.199** | **48.530±7.579** | **0.995±0.005** |
| PConv-Multi | 8.412±7.457 | 40.689±4.560 | 0.986±0.010 |
| GConv-Multi | 6.155±6.536 | 42.828±5.659 | 0.989±0.00 |
| GConv$_{Lap}$-Multi | **4.851±5.241** | **44.268±6.287** | **0.991±0.008** |

Similar to the LC experiments, for the HN dataset, the proposed GConv$_{Lap}$ model outperforms the other models with respect to the quality of the inpainted images. It should be noted that both LC and HN datasets were trained and tested under similar conditions, including the network architecture, and hyperparameters. Therefore, the reason that the range of the reported numerical values is different between the two datasets is related to the fact that the HN images entail less content and texture compared to the LC images. In addition to assessing the inpainting models with multimodal datasets, the models were trained and tested with single-modality images as well. In other words, for each of the LC and HN datasets, CT images and PET images were independently used to train and test the quality of the inpainting models (Tables 1.1. and 1.2. in Supplementary Materials). Similar to multimodal inpainting networks, even for the single modality images, GConv$_{Lap}$ outperformed the other models with a rather remarkable margin. To test the statistically significant differences between the performance of the GConv$_{Lap}$ model and the two other inpainting baselines, Wilcoxon signed rank test as a non-parametric method was applied to the calculated image quality metrics (see Table 1.3. in Supplementary Material).

Figure 5 demonstrates the qualitative comparisons between the functionality of the inpainting models in filling the random holes with meaningful patterns in the multimodal LC dataset. The irregular holes were randomly distributed over different locations on the image plane to learn the heterogeneous appearance of anatomical structures such as ribs, cardiac muscle, aorta, arteries, chest wall, etc.

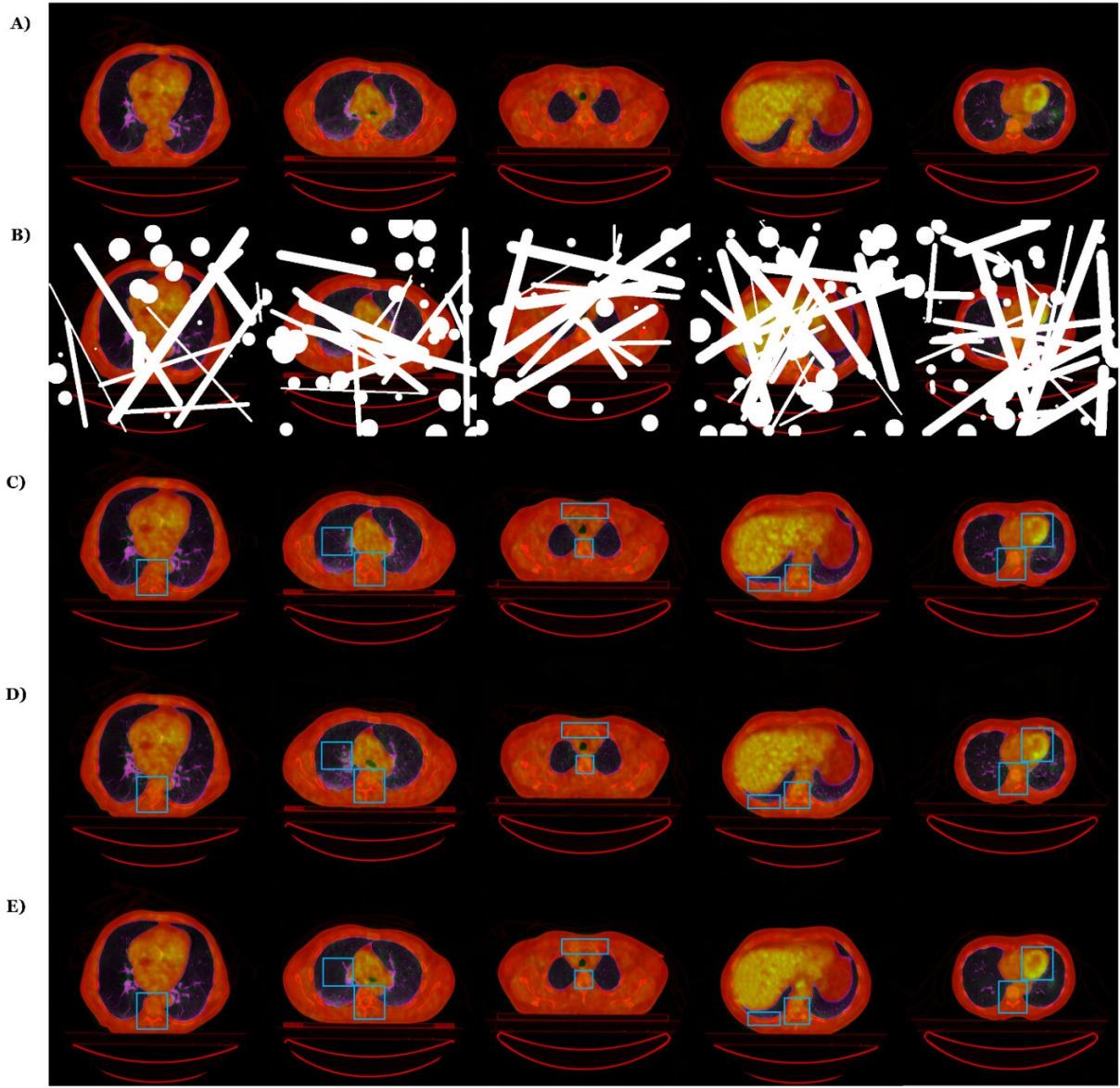

Figure 5. Qualitative comparisons of image inpainting performance. Row A: original PET-CT slices; row B: corrupted slices with random holes; row C: inpainted results by PConv model; row D: inpainted results by GConv model; and row E: inpainted results by the proposed GConvLap model. The proposed GConvLap model could replace the irregular holes with meaningful anatomical patterns and preserve the anatomical constraints far better than the other two methods. The blue bounding boxes highlighted the regions where the inpainted patterns by the proposed model are more meaningful anatomically than the other models.

For both the LC and HN datasets, quantitative values show that the performance of the proposed $GConv_{Lap}$ model is far better than the PConv model and slightly more accurate than the GConv model. Nonetheless, the capability of the proposed $GConv_{Lap}$ model in preserving the anatomical constraints is highlighted in Figure 5. In specific, while the PConv and GConv models filled the random holes with semantic image contents, they were not able to synthesize anatomically meaningful contents. From the qualitative comparisons between the anatomical regions highlighted with the blue boxes in Figure 5, it can be understood that the proposed $GConv_{Lap}$ model synthesized plausible image contents with highly realistic anatomical details. Therefore, both the image details and contextual patterns of the inpainted images synthesized by the proposed model are more similar to those of the original images, which in turn leads to reducing reconstruction errors.

## 3.2. Autoinpainting for tumor segmentation

The AutoPET and HECKTOR datasets were analyzed with a subject-wise 5-fold cross-validation strategy. In specific, for each fold, healthy slices of 80% of subjects were employed to train the inpainting models, and all the slices from the rest of the 20% of subjects were used to examine the autoinpainting pipeline for tumoral removal. Furthermore, the best-performing fold of the AutoPET models was used for the prediction of the internal LC dataset. The performance of the proposed autoinpainting pipeline for tumor segmentation is quantified by finding the agreement between the segmented volumes and the label masks. The autoinpainting pipeline was applied to all three inpainting models, followed by the same postprocessing steps for tumor segmentation. Tables 3, 4, and 5 represent the segmentation accuracy of the proposed autoinpainting strategy for studied LC and HN tumors.

Table 3 – Numerical results of tumor segmentation over the AutoPET LC dataset with the proposed autoinpainting method. The best-quantified metric for each of the model-data is marked in bold.

| Model-Data | Quantitative Metrics ($\mu \pm \sigma$) | | | |
|---|---|---|---|---|
| | [Dice] | [Precision] | [Recal] | Dice |
| PConv-CT | 0.388±0.183 | 0.408±0.214 | 0.386±0.162 | 0.357±0.103 |
| GConv-CT | 0.429±0.198 | 0.418±0.222 | 0.436±0.185 | 0.403±0.101 |
| GConv$_{Lap}$-CT | **0.452±0.202** | **0.437±0.225** | **0.490±0.186** | **0.426±0.082** |
| PConv-PET | 0.751±0.172 | 0.825±0.145 | 0.708±0.180 | 0.729±0.182 |
| GConv-PET | 0.761±0.163 | **0.826±0.146** | 0.720±0.175 | 0.740±0.168 |
| GConv$_{Lap}$-PET | **0.790±0.157** | 0.815±0.151 | **0.782±0.151** | **0.770±0.175** |
| PConv-Multi | 0.701±0.167 | **0.850±0.090** | 0.630±0.193 | 0.642±0.151 |
| GConv-Multi | 0.749±0.178 | 0.842±0.096 | 0.665±0.206 | 0.678±0.151 |
| GConv$_{Lap}$-Multi | **0.788±0.153** | 0.825±0.128 | **0.714±0.171** | **0.730±0.165** |

Table 4 – Numerical results of tumor segmentation over the internal LC dataset with the proposed autoinpainting method. The best-quantified metric for each of the model-data is marked in bold.

| Model-Data | Quantitative Metrics ($\mu \pm \sigma$) | | | |
|---|---|---|---|---|
| | [Dice] | [Precision] | [Recal] | Dice |
| PConv-CT | 0.382±0.157 | 0.408±0.186 | 0.389±0.151 | 0.353±0.111 |
| GConv-CT | 0.423±0.180 | 0.463±0.199 | 0.411±0.178 | 0.398±0.124 |
| GConv$_{Lap}$-CT | **0.442±0.176** | **0.482±0.192** | **0.426±0.176** | **0.410±0.134** |
| PConv-PET | 0.709±0.215 | 0.793±0.196 | 0.669±0.221 | 0.654±0.132 |
| GConv-PET | **0.750±0.176** | 0.792±0.192 | **0.747±0.189** | **0.690±0.184** |
| GConv$_{Lap}$-PET | 0.746±0.196 | **0.822±0.169** | 0.706±0.217 | 0.686±0.121 |
| PConv-Multi | 0.673±0.245 | 0.771±0.219 | 0.622±0.252 | 0.625±0.122 |
| GConv-Multi | 0.747±0.172 | 0.799±0.178 | 0.718±0.183 | 0.692±0.136 |
| GConv$_{Lap}$-Multi | **0.766±0.171** | **0.832±0.158** | **0.726±0.184** | **0.708±0.118** |

From Tables 3 and 4, we can observe that the segmentation accuracy achieved by the proposed GConv$_{Lap}$ model is remarkably higher than that of the PConv model, regardless of the type of input images. The same trend can be seen when comparing the GConv$_{Lap}$ model with the ordinary GConv model except for the case of inference on internal PET images where the GConv model slightly performs better than the GConv$_{Lap}$ model.

Table 5 – Numerical results of HN tumor segmentation with autoinpainting pipeline. The best-quantified metric for each of the model-data is marked in bold.

| Model-Data | Quantitative Metrics ($\mu \pm \sigma$) | | | |
|---|---|---|---|---|
| | [$Dice$] | [$Precision$] | [$Recal$] | $Dice$ |
| PConv-CT<br>GConv-CT<br>GConv$_{Lap}$-CT | NA | NA | NA | NA |
| PConv-PET<br>GConv-PET<br>GConv$_{Lap}$-PET | 0.686±0.105<br>0.721±0.183<br>**0.743±0.109** | 0.731±0.125<br>0.751±0.148<br>**0.772±0.113** | 0.669±0.162<br>0.701±0.214<br>**0.720±0.118** | 0.635±0.173<br>0.662±0.201<br>**0.674±0.149** |
| PConv-Multi<br>GConv-Multi<br>GConv$_{Lap}$-Multi | 0.693±0.192<br>0.714±0.107<br>**0.737±0.144** | 0.822±0.153<br>0.843±0.117<br>**0.866±0.105** | 0.629±0.162<br>0.642±0.146<br>**0.658±0.172** | 0.620±0.171<br>0.648±0.104<br>**0.669±0.089** |

Similar to the LC tumors, the segmentation accuracy of HN tumors achieved by the proposed GConv$_{Lap}$ outperformed the PConv model with a relatively large margin and performed more accurately than the ordinary GConv model on the PET and multimodal images. The appearance, textural distributions, and Hounsfield values of the HN tumors are very similar to those of the surrounding soft tissues (see Figure 1.1. in Supplementary Materials). Hence, the HN tumors in CT images cannot be inpainted by relying only on visible anatomical contrasts and not taking into account the mass effects. Therefore, none of the inpainting approaches is able to detect the deformation caused by the presence of HN tumors in CT slices. In this domain, it is worth mentioning that even the supervised segmentation models can hardly detect the HN tumors in full-resolution CT images (see Table 6). The proposed unsupervised autoinpainting pipeline was not able to detect the HN tumors in CT images; therefore, the notation of "NA" was used in the relevant rows of Table 5. Table 1.4. in Supplementary Materials shows the results of the applied Wilcoxon signed rank test on the Dice values achieved by the autoinpainting pipeline.

Figure 6 illustrates the capability of the proposed autoinpainting method in segmenting the tumors in multimodal images. Figure 1.2. in Supplementary Materials depicts a similar illustration for single modality images.

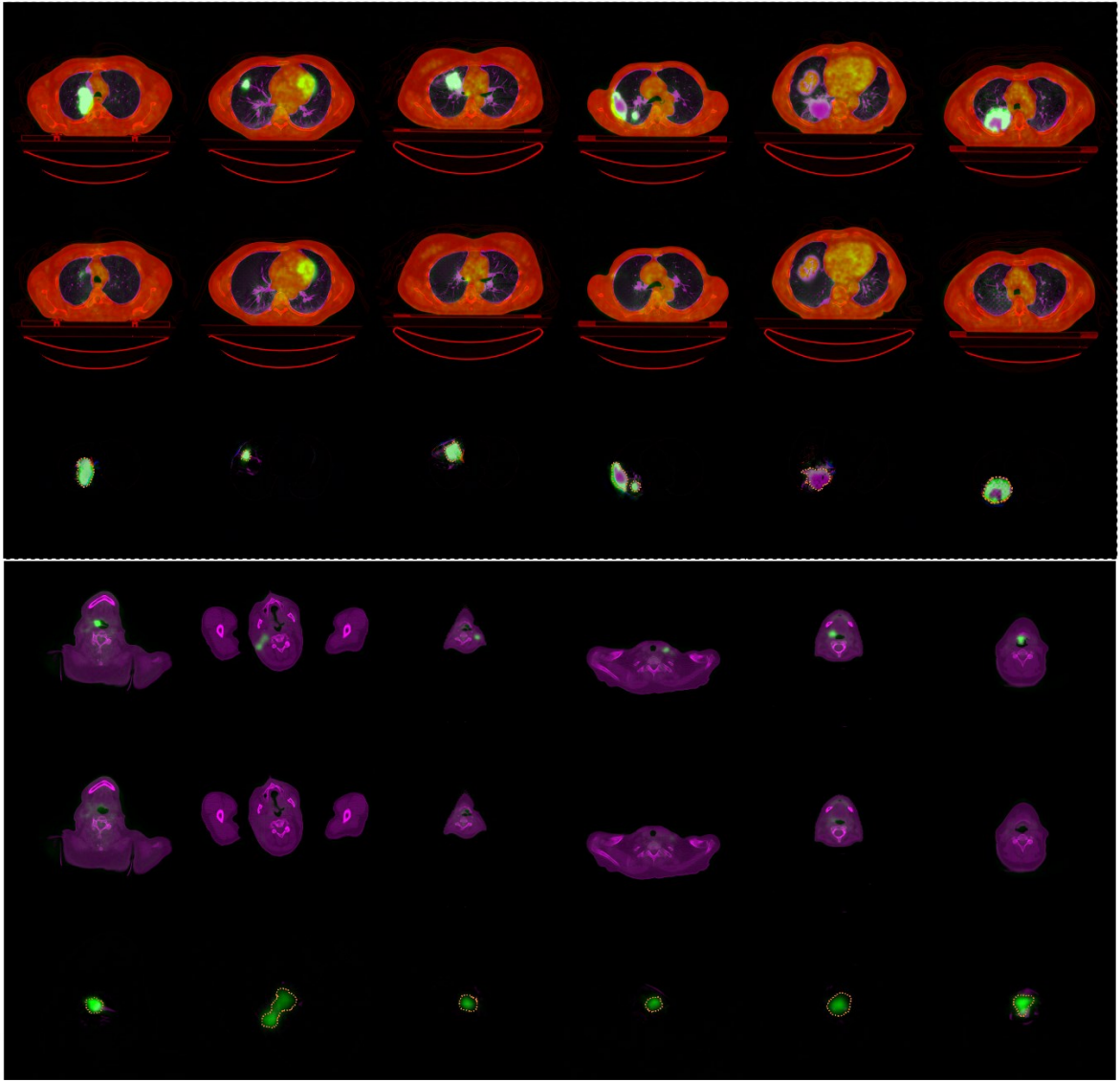

Figure 6. Visualization of the segmentation performance of the proposed autoinpainting pipeline in six exemplary cases. For each of the LC and HN images, the first row shows the original tumoral slices, the second row depicts the result of the proposed autoinpainting model, and the last row demonstrates the residuals between the two images. Please note that residual images were zoomed around the tumoral candidates to better visualize the qualitative comparison between the detected tumors and the ground truth (dashed orange contours).

There are certain cases in which the proposed unsupervised method faces some difficulties in segmenting the tumors. Figures 1.3. to 1.6. in Supplementary Materials depict different examples of challenging cases where the proposed pipeline failed to completely remove the tumors.

3.3. Supervised tumor segmentation

Table 6 presents the segmentation accuracy of the supervised nnU-Net model, which was trained with a 5-fold cross-validation resampling technique for each of the LC and HN tumors independently.

Table 6 – Numerical results of supervised segmentation accuracy achieved by the nn-UNet model. For each experiment, the differences in Dice scores w.r.t. the best unsupervised autoinpainting model are indicated in the parenthesis.

| Data-Tumor-Modality | Quantitative Metrics ($\mu \pm \sigma$) | | |
|---|---|---|---|
| | Dice | Precision | Recall |
| Internal-LC-CT | 0.707±0.224 (+0.297) | 0.762±0.238 | 0.713±0.258 |
| Internal-LC-PET | 0.802±0.177 (+0.112) | 0.802±0.231 | 0.854±0.174 |
| Internal-LC-Multi | 0.802±0.169 (+0.094) | 0.847±0.182 | 0.812±0.231 |
| AutoPET-LC-CT | 0.589±0.209 (+0.163) | 0.565±0.235 | 0.692±0.21 |
| AutoPET-LC-PET | 0.809±0.152 (+0.039) | 0.803±0.177 | 0.855±0.136 |
| AutoPET-LC-Multi | 0.818±0.141 (+0.088) | 0.812±0.160 | 0.854±0.145 |
| HECKTOR-HN-CT | 0.663±0.195 (NA) | 0.693±0.208 | 0.681±0.220 |
| HECKTOR-HN-PET | 0.697±0.174 (+0.023) | 0.751±0.188 | 0.701±0.205 |
| HECKTOR-HN-Multi | 0.753±0.150 (+0.084) | 0.793±0.162 | 0.758±0.190 |

Similar to the autoinpainting results, the supervised segmentation accuracy over the multimodal and PET images is higher than CT images for both LC and HN tumors. Moreover, integrating both modalities into the segmentation pipeline yielded the best results, which were even more accurate than PET images alone.

As was expected, the supervised models segment the tumors more accurately than the proposed unsupervised pipeline. However, carefully comparing the results, we can observe that the performance of the unsupervised autoinpainting models is not far behind the powerful supervised nnU-Net models for the cases of multimodal and PET images. For instance, the Dice scores achieved by the proposed GConv$_{Lap}$ model for multimodal LC tumors are 0.708 and 0.730 for the internal and AutoPET datasets. The Dice scores of the nnU-Net model of the same data are 0.802 and 0.818, respectively. The same competitive results can be observed for the HN tumor segmentations. In specific, the Dice scores achieved by the proposed autoinpainting method over the multimodal and PET images of the HN tumors are 0.669 and 0.674 while the supervised model resulted in Dice scores of 0.753 and 0.697, respectively. However, as was already described in section 2.6, comparing the differences between the supervised and unsupervised methods is not fair. In fact, the reason that the supervised nnU-Net model was examined is to estimate the optimal accuracy which can be achieved on the studied datasets.

### 3.4. Tumor segmentation with UAD methods

The segmentation accuracy of the employed UAD methods in multimodal images is presented in Tables 7, 8, and 9. Tables 1.5. to 1.9. in Supplementary Materials show similar evaluations for single modality images. In fact, eight conventional UAD models have been examined to benchmark the performance of the proposed unsupervised autoinpainting.

Table 7 – Segmentation accuracy of unsupervised anomaly detection models on multimodal images of AutoPET LC dataset. The metric of the best-performing benchmark model is marked in bold and the results of the proposed model are presented in italics type.

| Model | Quantitative Metrics ($\mu \pm \sigma$) | | | |
|---|---|---|---|---|
| | [Dice] | [Precision] | [Recall] | Dice |
| dAE | 0.232±0.109 | 0.227±0.141 | 0.291±0.111 | 0.213±0.053 |
| sAE | 0.147±0.059 | 0.163±0.100 | 0.165±0.054 | 0.135±0.044 |
| ceAE | **0.241±0.115** | **0.244±0.148** | 0.286±0.112 | **0.223±0.057** |
| VAE | 0.231±0.114 | 0.222±0.138 | 0.291±0.112 | 0.212±0.053 |
| ceVAE | 0.149±0.072 | 0.122±0.071 | 0.227±0.087 | 0.141±0.040 |
| GMVAE | 0.033±0.020 | 0.017±0.011 | **0.475±0.074** | 0.034±0.006 |
| F-AnoGAN | 0.128±0.079 | 0.093±0.074 | 0.361±0.148 | 0.116±0.022 |
| AAE | 0.215±0.115 | 0.215±0.150 | 0.288±0.123 | 0.195±0.047 |
| Autoinpainting | *0.788±0.153* | *0.825±0.128* | *0.714±0.171* | *0.730±0.165* |

Table 8 – Segmentation accuracy of unsupervised anomaly detection models on multimodal images of internal LC dataset. The metric of the best-performing benchmark model is marked in bold and the results of the proposed model are presented in italics type.

| Model | Quantitative Metrics ($\mu \pm \sigma$) | | | |
|---|---|---|---|---|
| | [Dice] | [Precision] | [Recall] | Dice |
| dAE | 0.305±0.122 | 0.270±0.132 | 0.405±0.147 | 0.285±0.068 |
| sAE | 0.097±0.047 | 0.064±0.038 | 0.249±0.072 | 0.094±0.030 |
| ceAE | **0.346±0.129** | **0.330±0.1464** | 0.407±0.144 | **0.314±0.078** |
| VAE | 0.311±0.132 | 0.271±0.142 | 0.421±0.158 | 0.282±0.068 |
| ceVAE | 0.254±0.109 | 0.228±0.126 | 0.320±0.119 | 0.242±0.069 |
| GMVAE | 0.023±0.016 | 0.012±0.008 | **0.583±0.117** | 0.023±0.004 |
| F-AnoGAN | 0.262±0.133 | 0.286±0.158 | 0.390±0.180 | 0.262±0.073 |
| AAE | 0.277±0.129 | 0.284±0.167 | 0.335±0.159 | 0.237±0.059 |
| Autoinpainting | *0.766±0.171* | *0.832±0.158* | *0.726±0.184* | *0.708±0.118* |

Comparing the numerical values of Tables 7, and 8 one can obviously observe that the proposed autoinpainting pipeline significantly outperformed all the UAD models on LC tumors. In specific, the best Dice scores in the UAD family were achieved by the ceAE model as 0.223 and 0.314 for multimodal AutoPET and internal datasets, respectively. However, these values are 50.7, and 39.4 percent inferior to the $GConv_{Lap}$ model (Dice=0.730, 0.708). The same trend can be observed for the single modality images when comparing the segmentation accuracy of the proposed autoinpainting model against the UAD models.

Table 9 – Segmentation accuracy of unsupervised anomaly detection models on multimodal images of HN tumors. The metric of the best-performing benchmark model is marked in bold and the results of the proposed model are presented in italics type.

| Model | Quantitative Metrics ($\mu \pm \sigma$) | | | |
|---|---|---|---|---|
| | [Dice] | [Precision] | [Recall] | Dice |
| dAE | 0.196±0.071 | 0.162±0.148 | 0.279±0.111 | 0.171±0.034 |
| sAE | 0.080±0.046 | 0.068±0.064 | 0.257±0.226 | 0.066±0.019 |
| ceAE | 0.238±0.072 | 0.167±0.149 | 0.274±0.105 | 0.211±0.034 |
| VAE | **0.269±0.086** | **0.196±0.179** | 0.397±0.104 | **0.241±0.033** |
| ceVAE | 0.189±0.082 | 0.156±0.163 | 0.318±0.176 | 0.179±0.023 |
| GMVAE | 0.049±0.028 | 0.026±0.016 | **0.479±0.090** | 0.049±0.008 |
| F-AnoGAN | 0.234±0.092 | 0.164±0.160 | 0.360±0.103 | 0.191±0.025 |
| AAE | 0.176±0.099 | 0.139±0.195 | 0.287±0.130 | 0.162±0.028 |
| Autoinpainting | *0.737±0.144* | *0.866±0.105* | *0.658±0.172* | *0.669±0.089* |

The UAD models faced serious difficulties to deal with even more challenging HN tumors. In other words, while the proposed $GConv_{Lap}$ model could achieve a segmentation accuracy of 0.669 in multimodal HN tumors, the examined UAD models barely obtained a Dice score of 0.241. Similar behavior was observed with PET images, where the proposed autoinpainting model outperformed the UAD models significantly. However, it should be noted that both UAD models and the proposed autoinpainting method failed to segment the HN tumors in CT images.

Figure 7 visualizes a qualitative comparison between the proposed autoinpainting method and the employed UAD models. Such comparisons signify the superiority of the proposed unsupervised autoinpainting approach over the conventional UAD models. In fact, the ability of the $GConv_{Lap}$ model to reconstruct high-resolution images by preserving the anatomical constraints on one side and its potential to detect and remove the tumors without corrupting the remaining anatomical structures on the other side boost the performance of the autoinpainting approach. On the other hand, the UAD models can neither preserve the anatomical constraints nor completely replace the tumors with healthy tissues. Figures 1.7. and 1.8. in Supplementary Materials show the same concept for the PET-CT images of HN tumors and CT images of LC tumors.

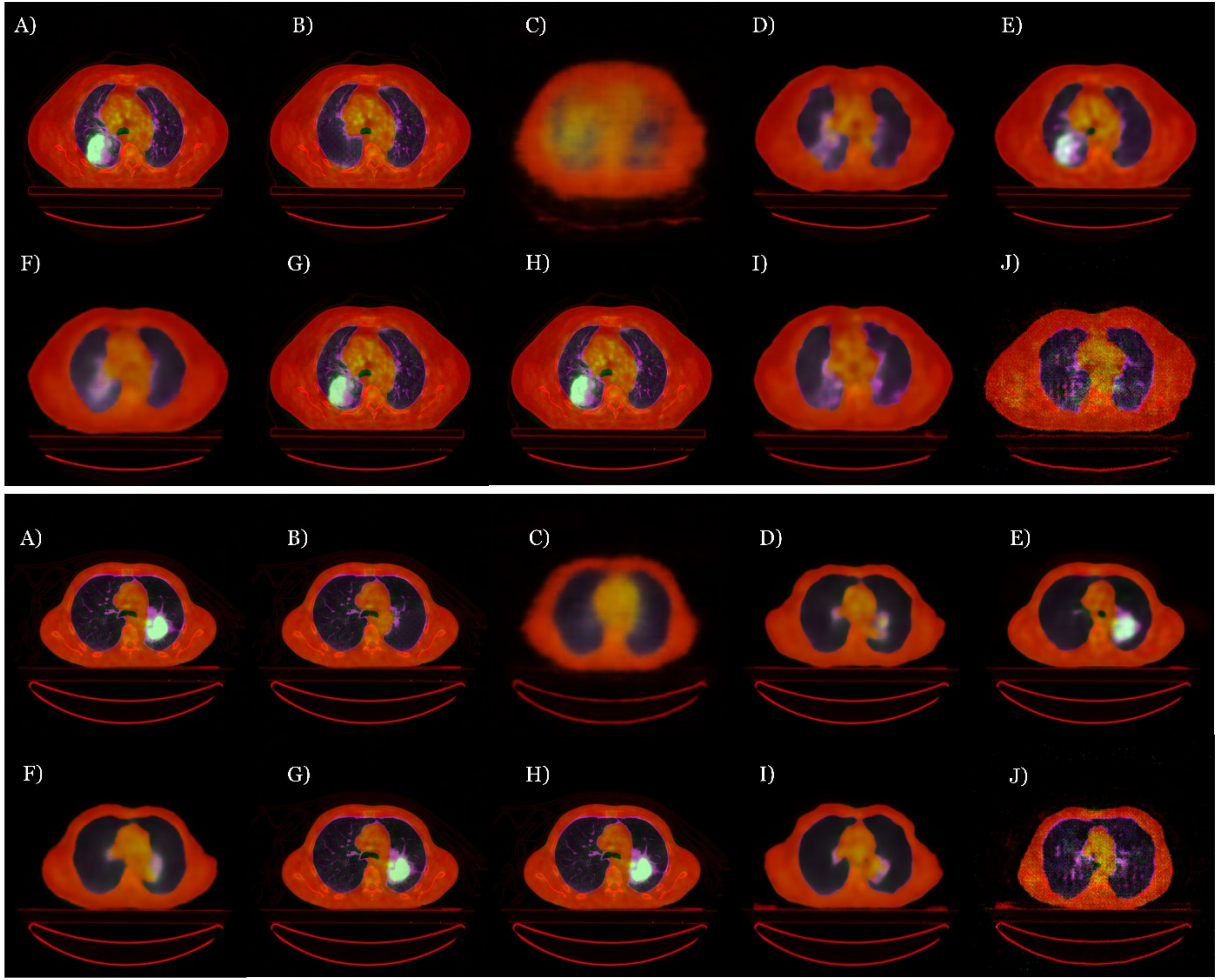

Figure 7. Qualitatively comparing the performance of the proposed autoinpainting pipeline against eight UAD models in learning the appearance of healthy lungs for an exemplary patient. Each set of images consists of A) original tumoral slice, B) proposed autoinpainted image, C) adversarial autoencoder result, D) dense autoencoder result, E) spatial autoencoder result, F) variational autoencoder result, G) context-encoding variational autoencoder result, H) Gaussian mixture variational autoencoder result, I) context-encoding autoencoder result, and J) Fast-Anomaly GAN.

## 4. Discussion and conclusion

The detection and segmentation of tumors in medical images support a series of important clinical tasks, including diagnosis, prognosis, treatment, and surgery planning. The development of accurate computerized methods for automatic tumor segmentation has become a major endeavor in medical image analysis communities. Recent advances in deep learning-based methods have led to the development of robust models which could achieve even expert-level performance in some applications. However, most of the developed models depend on an explicitly defined target class for their supervised training procedures. This dependency, in general, increases the sensitivity to the quality and quantity of the available labeled data, which in turn limits the generalization power of the models over the unseen and/or underpresented classes. Recently, to overcome the necessity of expensive labeled data, UAD methods have emerged as promising tools to detect pathologies from arbitrary types. These methods aim to resemble how radiologists examine imaging scans. In fact, expert radiologists are trained to learn the appearance of healthy anatomical regions. Therefore, they do not need data with pixel-level annotations because they can detect arbitrary abnormalities as outliers with respect to healthy anatomies (Baur et al., 2021; Pinaya et al., 2022). However, one of the limitations of conventional UAD models is that they hardly learn the appearance of healthy anatomical structures with fine-grained details. In specific, they often tend to learn a general representation of anatomical structures without preserving the details of anatomical constraints. The main objective of this study has been focused on developing an autoinpainting model to segment the tumors by generating high-resolution medical images without the tumors while preserving the anatomical details in the process of representation learning. To this end, we propose a robust image inpainting model, $GConv_{Lap}$, which is capable of capturing the appearance of normal anatomies and can synthesize high-resolution medical images by preserving the fine-grained anatomical details. In fact, one emphasis of this work has been to improve the performance of the conventional inpainting models for synthesizing medical images by preserving textural and anatomical details as much as possible. This inpainting model was trained with healthy image slices to model the characteristics of healthy anatomies by learning to fill the irregular random holes with anatomical and visually meaningful patterns. Then, an autoinpainting pipeline was developed to automatically inpaint the tumoral regions and synthesize high-quality tumor-free images. In fact, we hypothesized that the well-trained inpainting model would replace the tumoral tissues with the characteristics of already learned healthy structures and leave the healthy parts of the images intact. Therefore, the differences between the original tumoral images and the synthesized inpainted images can be used to segment the tumoral regions.

    The conventional AE-based models are often trained by optimizing per-pixel loss functions that tend to reconstruct blurry images. One potential approach is to modify the objective function in order to improve the quality of the reconstructed images. Therefore, more advanced loss functions such as perceptual loss and style loss can potentially increase the conceptual and textural quality of the generated images. However, integrating these objective functions into the conventional representation learning models such as AE models would degrade their ability to learn the latent characteristics of the healthy anatomies. In other words, such modified models tend to learn a wide range of image-based details and hardly can discriminate normal structures from anomalies. In fact, such fortified objective functions increase the risk of model overfitting with respect to representation learning tasks. However, limiting the functionality of convolutional operators within image subregions can regularize the learning process of representation learning models to avoid the overfitting problem. In particular, while the powerful objective function is prone to overfit on the details of anatomical structures, localizing the functionality of convolutional operators can potentially counteract this unwanted behavior. Accordingly, considering the functionality of the GConv operators, they can be a perfect choice for this problem as they deal with local convolutions instead of ordinary global convolutions. As a result, the representation learning process in this study was turned from conventional AE and GAN-based models into an image inpainting problem. In practice, leveraging the inpainting model with multi-term objective function as an optimization algorithm and GConv operator as localized convolutional backbones could successfully enforce the model to synthesize the high fidelity realistic-looking medical images while preserving the anatomical constraints regardless of the imaging modality. In practice, integrating the GConv operator into a U-Net-like architecture optimized by a multi-term objective function that is fortified by the Laplacian loss could successfully improve the quality of the inpainted images. In particular, the quantified metrics of Tables 1, 2, and Tables 1.1. and 1.2. in Supplementary Materials verify the superiority of the proposed $GConv_{Lap}$ model. The learnable soft mask updating procedure of the $GConv_{Lap}$ operator heuristically updates the invalid pixels, which leads to reconstructing images with more fidelities compared to the hard-gating rules embedded in the PConv operator. This effect is more evident by comparing the quality of the inpainted images by the three models when multimodal PET-CT images were used (such as Figure 5). Besides that, employing an encoder-decoder network architecture with skip connections could propagate the detailed color and textural information to the decoding path and fill the hole boundaries with smooth patterns. In addition, leveraging the objective function with Laplacian loss was a beneficial strategy to preserve both high and low-frequency patterns and synthesize images with fine-grained details as much as possible. In fact, one of the limitations of the PConv and GConv models is their difficulties in preserving the anatomical constraints, especially in the edges, such as transitions between soft and hard tissues or sharp intensity changes within soft tissues. As can be seen in Figure 5, both PConv and GConv were unable to reconstruct meaningful anatomical details, while the proposed $GConv_{Lap}$ model synthesized images with the highest similarity with respect to the original image slices regardless of the level of corruption applied to the images. Such qualitative comparison is consistent with the numerical values reported in Tables 1 and 2, which point to the advantages of the proposed inpainting model. Therefore, the proposed $GConv_{Lap}$ model can produce high-resolution images with the least level of anatomical distortions and false positives which makes it a suitable choice to be used as an UAD model.

    The proposed autoinpainting pipeline for tumor segmentation yielded interesting results in the context of unsupervised segmentation. In fact, the segmentation accuracy of the proposed unsupervised pipeline was not far behind the performance of the supervised nnU-Net model when the PET images were included either as multimodal or single-modality image data. In specific, the performance of the examined supervised models over the multimodal images of

internal LC, AutoPET LC, and HECKTOR HN datasets are, respectively, 9.4, 8.8, and 8.4% higher than the proposed unsupervised approach. This can be explained by the fact that the hyper signal intensity in PET images caused by tumoral uptakes facilitates tumor localization. Nevertheless, this should be noted that not all the hyperactive regions are related to cancerous tissues. In other words, other healthy tissues such as the cardiac muscle, among others take up high levels of FDG and often appear with hyperintensity patterns. Therefore, localization and segmentation of tumors in PET and multimodal PET-CT images is not a trivial task. In addition, the capabilities of the proposed inpainting model were not limited only to hyperintensity signals of PET images, as the pipeline could detect and inpaint the challenging LC tumors in CT images as well. Highly similar visual attributes of LC tumors with respect to the surrounding soft tissues make them challenging for segmentation models, even for the supervised ones. Nevertheless, the proposed autoinpainting strategy could inpaint and segment the challenging cases and lead to rather acceptable results in the context of UAD. Comparing the segmentation accuracy of $GConv_{Lap}$ model with PConv and the ordinary GConv model within the proposed autoinpainting framework signifies the superiority of the proposed inpainting model (Tables 3, 4, and 5). In particular, the advantage of GConv operator over the PConv module on one side and the ability of the proposed model to preserve the anatomical constraints on the other side lead to inpainting the tumoral regions while retaining the healthy structures intact. Therefore, tumoral tissues were removed by the proposed autoinpainting while the healthy structures were not manipulated, which resulted in remarkably fewer false positives. In this context, segmentation accuracies of LC tumors in CT images achieved by the supervised nnU-Net were 29.7 and 16.3% higher than the proposed method for the internal and AutoPET LC datasets. As expected, the tumor segmentation in PET images resulted in more accurate results than in CT images. In specific, while Dice scores of 0.410 and 0.426 were achieved by the $GConv_{Lap}$ model for studied LC tumor segmentation in CT images, this metric was improved to 0.686 and 0.770 for the PET images on the same dataset. Yet, it should be noted that the proposed unsupervised model failed to detect the HN tumors in CT images while the 3D nnU-Net model already achieved a segmentation accuracy of 0.663 in terms of the Dice metric. As can be seen in Figure 1.1. in Supplementary Materials, the lack of intensity and the textural contrast between the tumors and nearby soft tissues prevent the autoinpainting method from recognizing the tumoral regions as anomalies. In fact, we chose the HN tumor segmentation task as a challenging problem to highlight the limitations of UAD methods in general and the proposed method, in specific. Such a limitation can be observed in the LC dataset as well when the lung collapses or the tumors appear in the middle of soft tissues (Figure 1.3. in Supplementary Materials). Nevertheless, analyzing the multimodal PET-CT images could improve the segmentation accuracy for both the LC and HN tumors.

Comparing the segmentation accuracy of the proposed pipeline against the conventional UAD methods highlights the great potential of the autoinpainting model. Numerically, the best Dice achieved by the examined UAD models are 0.314, 0.223, and 0.241 for multimodal images of internal, AutoPET, and HECKTOR datasets, respectively, which are 39.4, 50.7, and 42.8 percent inferior to the corresponding Dice metrics achieved by the proposed $GConv_{Lap}$ model. In fact, the UAD models could neither reconstruct healthy images from tumoral slices nor preserve anatomical structures. In other words, they either removed the tumors and synthesized new images with meaningless anatomical structures or preserved the anatomical structures but could not remove the tumors. It should be emphasized that even when the UAD models managed to remove the tumors, they reconstruct images with severe anatomical distortions, which resulted in a high rate of false positives. This challenges the underlying hypothesis of UAD models, which aim to model the distribution of healthy data. Carefully examining the images (Figure 7 and Figures 1.7. and 1.8. in Supplementary Materials) generated by the best performing UAD models such as VAE, ceVAE, and F-AnoGAN, one can deduce that such models reconstructed texture-free images which do not hold meaningful anatomical details. Therefore, the tumors can be partially detected from the residual images only because there are no meaningful textures within the reconstructed images. Such a major limitation of the current UAD methods was highlighted in a recent study (Meissen et al., 2021) in which the authors showed that even simple image processing algorithms as thresholding can yield competitive results to those of the UAD models. Other types of UAD methods aim to detect the anomalies but not directly from the residual maps between the original and the reconstructed images (Dey and Hong, 2021; van Hespen et al., 2021); therefore, such models do not aim to produce high-quality anomaly-free images either. In contrast to these methods, the proposed autoinpainting-based anomaly detection pipeline can capture the normal anatomies and generate high-resolution anomaly-free images by retaining fine-grained anatomical details.

The reason for choosing a circular shape window to sweep the images is based on the fact that a circle can fully cover the tumoral regions regardless of the irregularity of the tumor morphology and the healthy pixels between the tumoral borders and the circle boundaries help the inpainting model to fill the circular hole. In this context, while tumors appear with a wide range of sizes, the strategy of adaptively changing the size of the moving circle was beneficial to detect and inpaint both small and large-size tumors. Last but not least, the postprocessing pipeline contains a size-based thresholding step to avoid detecting small deviations as tumoral candidate regions. This thresholding step could potentially ignore the presence of tiny anomalies; however, the goal of this study has focused on segmenting clinically relevant tumors, not early-stage tiny nodules.

Finally, despite the efficacy of the proposed autoinapainting-based UAD model for segmenting tumors in multimodal and single-modal images, there exist some limitations within the proposed pipeline, which will worth investigating in future studies. In particular, the proposed sliding window for sweeping different coordinates of the images is an exhaustive strategy. Roughly clustering the candidate regions followed by the proposed autoinpaining method can reduce the computational time. Furthermore, extending the 2D autoinpainting pipeline into a 3D approach requires the development of a robust 3D inpainting model, which may further improve the accuracy of inpainting by incorporating the volumetric contexts. Also, if using 3D information, a dataset of healthy volunteers will be needed which is not available publicly at this time.

While the unsupervised segmentation methods aim to overcome the disadvantages of supervised models, the current UAD models have not been robust enough to yield as accurate results as supervised models. In this study, an inpainting-based UAD method was proposed to segment the LC and HN tumors in multimodal and single-modal images. To the best knowledge of the author, it has been the first attempt to segment such challenging tumors with unsupervised methods. The quantitative results show the potential of the proposed pipeline with superior performance over the conventional UAD models.

## 5. Acknowledgment

This study was supported by the Swedish Childhood Cancer Foundation (grant no. MT2019-0019), the Swedish innovation agency Vinnova (grant no. 2017-01247), the Swedish Research Council (VR) (grant no. 2018-04375) and the German Ministry of Education and Research (BMBF) (grant no. 13GW0357 A-C). We also thank Stockholm Medical Image Laboratory and Education (SMILE) for giving us access to their Nvidia DGX-1 server.

## 6. Declaration of generative AI and AI-assisted technologies in the writing process

This manuscript has been written and prepared without employing any type of AI or AI-assisted technology.

# AutoPaint: Suplementary Material

## 1) Supplementary Tables and Figures

Table 1.1. The numerical comparison between the performance of inpainting models trained on AutoPET and tested on internal LC dataset for single modality images

| Model-Data | Quantitative Metrics ($\mu \pm \sigma$) | | |
| --- | --- | --- | --- |
| | MSE | PSNR | SSIM |
| Pconv-CT | 109.883±63.296 | 28.572±2.966 | 0.921±0.034 |
| Gconv-CT | 67.955±51.888 | 31.770±5.161 | 0.943±0.033 |
| Gconv$_{Lap}$-CT | 62.061±47.406 | 32.096±5.011 | 0.949±0.030 |
| Pconv-PET | 16.336±17.698 | 38.382±4.494 | 0.980±0.012 |
| Gconv-PET | 16.040±21.927 | 39.351±6.115 | 0.981±0.013 |
| Gconv$_{Lap}$-PET | 15.668±19.547 | 39.312±6.112 | 0.982±0.012 |

Table 1.2. The numerical comparison between the performance of inpainting models trained and tested on HN single modality images

| Model-Data | Quantitative Metrics ($\mu \pm \sigma$) | | |
| --- | --- | --- | --- |
| | MSE | PSNR | SSIM |
| Pconv-CT | 9.533±9.009 | 40.429±5.066 | 0.985±0.013 |
| Gconv-CT | 6.815±6.944 | 42.757±6.414 | 0.989±0.011 |
| Gconv$_{Lap}$-CT | 5.542±6.686 | 44.072±6.917 | 0.991±0.009 |
| Pconv-PET | 6.073±7.484 | 47.142±7.363 | 0.993±0.007 |
| Gconv-PET | 5.852±17.400 | 48.533±6.909 | 0.993±0.008 |
| Gconv$_{Lap}$-PET | 3.413±10.395 | 50.123±9.312 | 0.995±0.006 |

Table 1.3. Statistical comparison of image quality metrics between the three inpainting models using the Wilcoxon signed rank test

| Data-Tumor-Modality | Models | p-value | | |
| --- | --- | --- | --- | --- |
| | | MSE | PSNR | SSIM |
| Internal-LC-CT | Pconv vs. Gconv | < 0.0001 | < 0.0001 | < 0.0001 |
| | Pconv vs. Gconv$_{Lap}$ | < 0.0001 | < 0.0001 | < 0.0001 |
| | Gconv vs. Gconv$_{Lap}$ | 0.665 | 0.430 | 0.0007 |
| Internal-LC-PET | Pconv vs. Gconv | 0.516 | 0.005 | 0.0005 |
| | Pconv vs. Gconv$_{Lap}$ | < 0.0001 | < 0.0001 | < 0.0001 |
| | Gconv vs. Gconv$_{Lap}$ | < 0.0001 | < 0.0001 | < 0.0001 |
| Internal-LC-Multi | Pconv vs. Gconv | < 0.0001 | < 0.0001 | < 0.0001 |
| | Pconv vs. Gconv$_{Lap}$ | < 0.0001 | < 0.0001 | < 0.0001 |
| | Gconv vs. Gconv$_{Lap}$ | 0.001 | 0.0139 | < 0.0001 |
| HECKTOR-HN-CT | Pconv vs. Gconv | < 0.0001 | < 0.0001 | < 0.0001 |
| | Pconv vs. Gconv$_{Lap}$ | < 0.0001 | < 0.0001 | < 0.0001 |
| | Gconv vs. Gconv$_{Lap}$ | < 0.0001 | < 0.0001 | < 0.0001 |
| HECKTOR-HN-PET | Pconv vs. Gconv | < 0.0001 | < 0.0001 | < 0.0001 |
| | Pconv vs. Gconv$_{Lap}$ | < 0.0001 | < 0.0001 | < 0.0001 |
| | Gconv vs. Gconv$_{Lap}$ | < 0.0001 | < 0.0001 | < 0.0001 |
| HECKTOR-HN-Multi | Pconv vs. Gconv | < 0.0001 | < 0.0001 | < 0.0001 |
| | Pconv vs. Gconv$_{Lap}$ | < 0.0001 | < 0.0001 | < 0.0001 |
| | Gconv vs. Gconv$_{Lap}$ | < 0.0001 | < 0.0001 | < 0.0001 |

Table 1.4. Statistical comparison of the achieved Dice scores by the autoinpainting method applied to the three inpainting models

| Data-Tumor-Modality | Models | p-value |
| --- | --- | --- |
| Internal-LC-CT | Pconv vs. Gconv | < 0.0001 |
| | Pconv vs. Gconv$_{Lap}$ | < 0.0001 |

| | Gconv vs. Gconv$_{Lap}$ | < 0.0001 |
|---|---|---|
| Internal-LC-PET | Pconv vs. Gconv | < 0.0001 |
| | Pconv vs. Gconv$_{Lap}$ | < 0.0001 |
| | Gconv vs. Gconv$_{Lap}$ | 0.083 |
| Internal-LC-Multi | Pconv vs. Gconv | < 0.0001 |
| | Pconv vs. Gconv$_{Lap}$ | < 0.0001 |
| | Gconv vs. Gconv$_{Lap}$ | < 0.0001 |
| AutoPET-LC-CT | Pconv vs. Gconv | < 0.0001 |
| | Pconv vs. Gconv$_{Lap}$ | < 0.0001 |
| | Gconv vs. Gconv$_{Lap}$ | < 0.0001 |
| AutoPET-LC-PET | Pconv vs. Gconv | < 0.0001 |
| | Pconv vs. Gconv$_{Lap}$ | < 0.0001 |
| | Gconv vs. Gconv$_{Lap}$ | < 0.0001 |
| AutoPET-LC-Multi | Pconv vs. Gconv | < 0.0001 |
| | Pconv vs. Gconv$_{Lap}$ | < 0.0001 |
| | Gconv vs. Gconv$_{Lap}$ | < 0.0001 |
| HECKTOR-HN-PET | Pconv vs. Gconv | < 0.0001 |
| | Pconv vs. Gconv$_{Lap}$ | < 0.0001 |
| | Gconv vs. Gconv$_{Lap}$ | 0.019 |
| HECKTOR-HN-Multi | Pconv vs. Gconv | < 0.0001 |
| | Pconv vs. Gconv$_{Lap}$ | < 0.0001 |
| | Gconv vs. Gconv$_{Lap}$ | < 0.0001 |

Table 1.5. Segmentation accuracy of unsupervised anomaly detection models on CT images of internal LC tumors

| Model | Quantitative Metrics ($\mu \pm \sigma$) | | | |
|---|---|---|---|---|
| | [$Dice$] | [$Precision$] | [$Recal$] | $Dice$ |
| dAE | 0.225+0.123 | 0.200+0.127 | 0.302+0.126 | 0.200+0.069 |
| sAE | 0.022+0.013 | 0.013+0.008 | 0.134+0.062 | 0.020+0.007 |
| ceAE | 0.221+0.123 | 0.201+0.141 | 0.295+0.122 | 0.198+0.068 |
| VAE | 0.210+0.114 | 0.182+0.121 | 0.287+0.107 | 0.190+0.066 |
| ceVAE | 0.102+0.059 | 0.072+0.050 | 0.220+0.118 | 0.094+0.029 |
| GMVAE | 0.013+0.007 | 0.006+0.003 | 0.519+0.108 | 0.013+0.002 |
| F-AnoGAN | 0.133+0.088 | 0.087+0.068 | 0.378+0.170 | 0.120+0.037 |
| AAE | 0.083+0.072 | 0.053+0.055 | 0.422+0.162 | 0.062+0.021 |

Table 1.6. Segmentation accuracy of unsupervised anomaly detection models on CT images of AutoPET LC tumors

| Model | Quantitative Metrics ($\mu \pm \sigma$) | | | |
|---|---|---|---|---|
| | [$Dice$] | [$Precision$] | [$Recal$] | $Dice$ |
| dAE | 0.197+0.122 | 0.184+0.136 | 0.257+0.119 | 0.178+0.060 |
| sAE | 0.062+0.037 | 0.047+0.034 | 0.134+0.076 | 0.055+0.022 |
| ceAE | 0.203+0.120 | 0.195+0.148 | 0.268+0.114 | 0.180+0.065 |
| VAE | 0.193+0.114 | 0.164+0.115 | 0.283+0.122 | 0.177+0.060 |
| ceVAE | 0.097+0.057 | 0.070+0.049 | 0.220+0.107 | 0.089+0.024 |
| GMVAE | 0.023+0.015 | 0.012+0.008 | 0.447+0.102 | 0.024+0.004 |
| F-AnoGAN | 0.2013+0.152 | 0.183+0.152 | 0.378+0.154 | 0.193+0.045 |
| AAE | 0.137+0.101 | 0.119+0.121 | 0.249+0.107 | 0.111+0.040 |

Table 1.7. Segmentation accuracy of unsupervised anomaly detection models on PET images of internal LC tumors

| Model | Quantitative Metrics ($\mu \pm \sigma$) | | | |
|---|---|---|---|---|
| | [$Dice$] | [$Precision$] | [$Recal$] | $Dice$ |
| dAE | 0.583+0.160 | 0.658+0.184 | 0.533+0.159 | 0.548+0.189 |
| sAE | 0.673+0.256 | 0.684+0.275 | 0.727+0.201 | 0.603+0.197 |
| ceAE | 0.564+0.142 | 0.641+0.162 | 0.515+0.145 | 0.530+0.187 |
| VAE | 0.710+0.173 | 0.752+0.190 | 0.684+0.174 | 0.664+0.223 |
| ceVAE | 0.478+0.161 | 0.525+0.205 | 0.473+0.155 | 0.454+0.098 |
| GMVAE | 0.026+0.017 | 0.013+0.009 | 0.653+0.094 | 0.026+0.004 |

| Model | | | | |
|---|---|---|---|---|
| F-AnoGAN | 0.683+0.199 | 0.745+0.215 | 0.643+0.197 | 0.634+0.184 |
| AAE | 0.393+0.145 | 0.525+0.257 | 0.324+0.246 | 0.343+0.203 |

Table 1.8. Segmentation accuracy of unsupervised anomaly detection models on PET images of AutoPET LC tumors

| Model | Quantitative Metrics ($\mu\pm\sigma$) | | | |
|---|---|---|---|---|
| | $[Dice]$ | $[Precision]$ | $[Recal]$ | $Dice$ |
| dAE | 0.534±0.143 | 0.493±0.201 | 0.594±0.241 | 0.493±0.093 |
| sAE | 0.372±0.249 | 0.324±0.216 | 0.394±0.149 | 0.350±0.104 |
| ceAE | 0.590±0.205 | 0.594±0.093 | 0.638±0.142 | 0.553±0.152 |
| VAE | 0.641±0.231 | 0.629±0.124 | 0.692±0.104 | 0.612±0.091 |
| ceVAE | 0.392±0.158 | 0.382±0.251 | 0.424±0.207 | 0.363±0.162 |
| GMVAE | 0.048±0.032 | 0.019±0.032 | 0.376±0.126 | 0.042±0.031 |
| F-AnoGAN | 0.693±0.142 | 0.703±0.134 | 0.729±0.123 | 0.652±0.174 |
| AAE | 0.355±0.281 | 0.341±0.182 | 0.389±0.248 | 0.319±0.173 |

Table 1.9. Segmentation accuracy of unsupervised anomaly detection models on PET images of HN tumors

| Model | Quantitative Metrics ($\mu\pm\sigma$) | | | |
|---|---|---|---|---|
| | $[Dice]$ | $[Precision]$ | $[Recal]$ | $Dice$ |
| dAE | 0.390±0.157 | 0.477±0.214 | 0.359±0.138 | 0.330±0.104 |
| sAE | 0.158±0.109 | 0.371±0.289 | 0.149±0.137 | 0.137±0.038 |
| ceAE | 0.402±0.153 | 0.481±0.211 | 0.376±0.129 | 0.337±0.107 |
| VAE | 0.437±0.207 | 0.471±0.240 | 0.455±0.174 | 0.391±0.113 |
| ceVAE | 0.271±0.180 | 0.469±0.308 | 0.339±0.229 | 0.260±0.050 |
| GMVAE | 0.064±0.037 | 0.033±0.021 | 0.756±0.120 | 0.064±0.011 |
| F-AnoGAN | 0.426±0.203 | 0.461±0.239 | 0.417±0.171 | 0.388±0.108 |
| AAE | 0.413±0.200 | 0.406±0.237 | 0.453±0.157 | 0.361±0.116 |

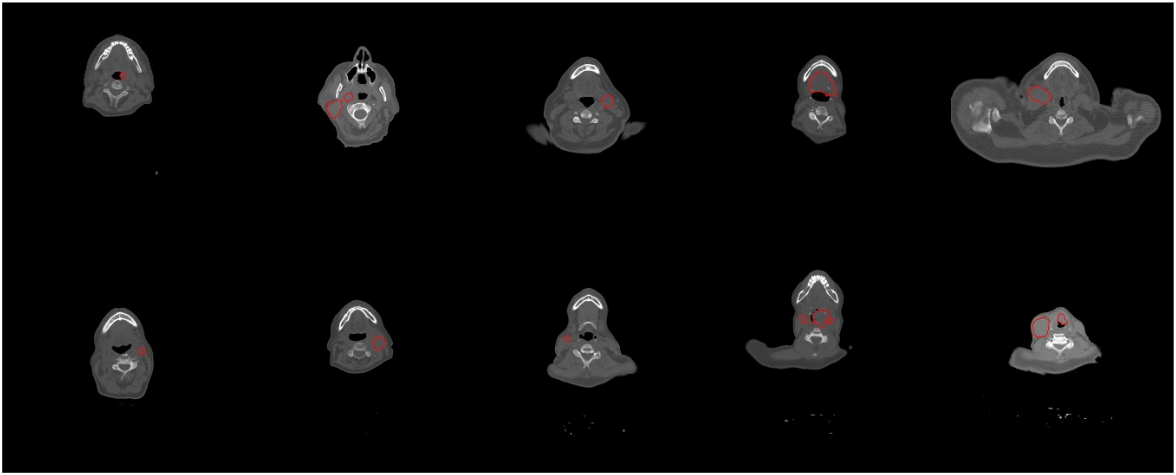

Figure 1.1. Examples of the appearance of HN tumors, highlighted in red contours, in CT images. The presence of tumors among the densely connected soft tissues with a similar range of Hounsfield values makes the segmentation of HN tumors a challenging problem.

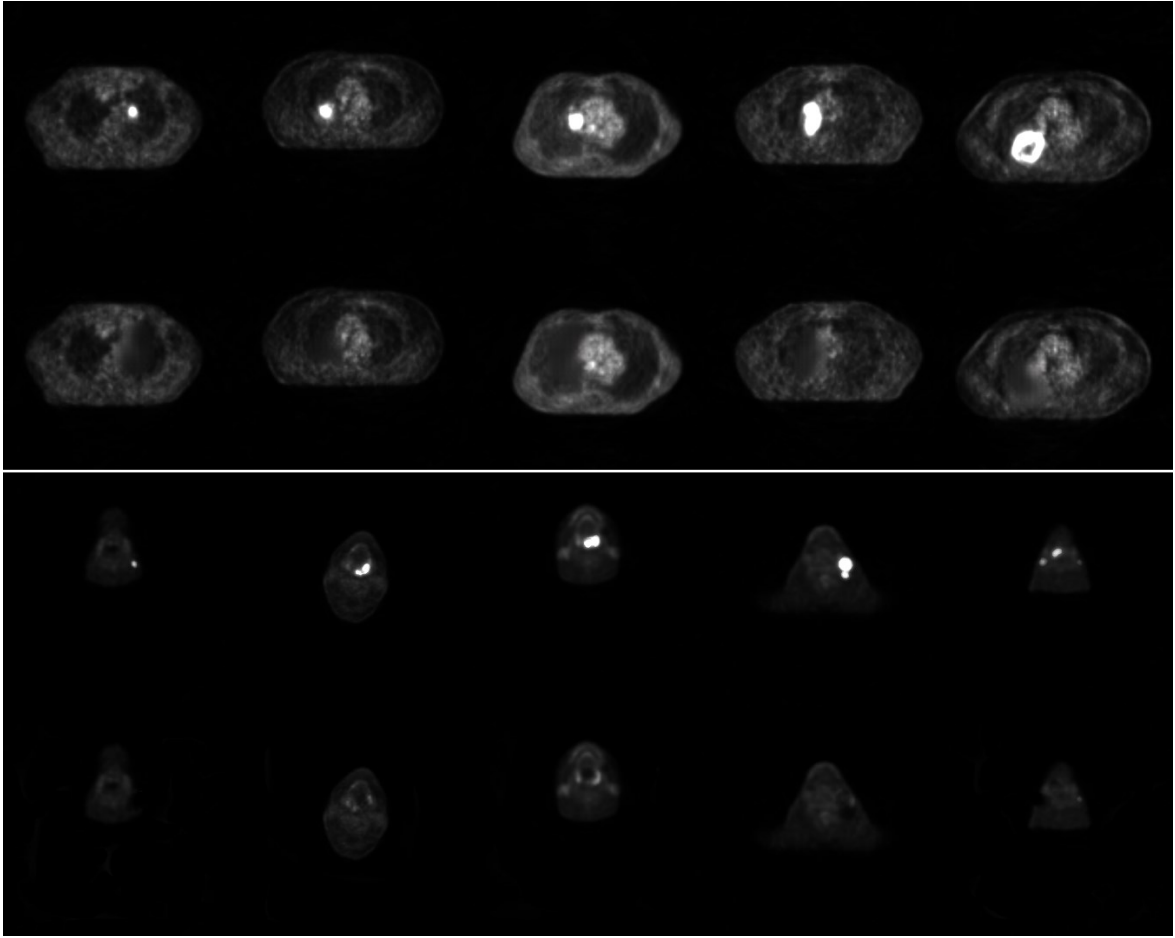

Figure 1.2. Visualization of the segmentation performance of the proposed autoinpainting pipeline for single-modality PET images. For each of the LC and HN images, the first row shows the original tumoral slices, and the second row depicts the result of the proposed autoinpainting model where the tumors were replaced by healthy tissues and fake tumor-free images were synthesized.

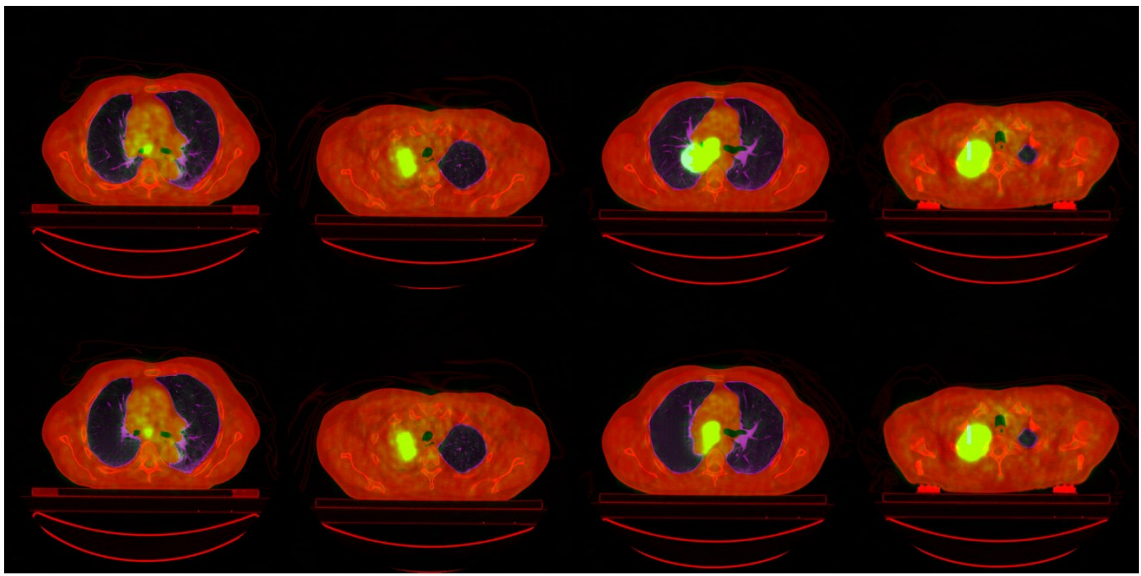

Figure 1.3. Examples of challenging LC tumors in multimodal PET-CT images where the proposed autoinpaiting pipeline either failed to detect the tumors or could only partially remove the tumors. The first row shows the original tumoral slices, and the second row depicts the results of autoinpainting pipeline

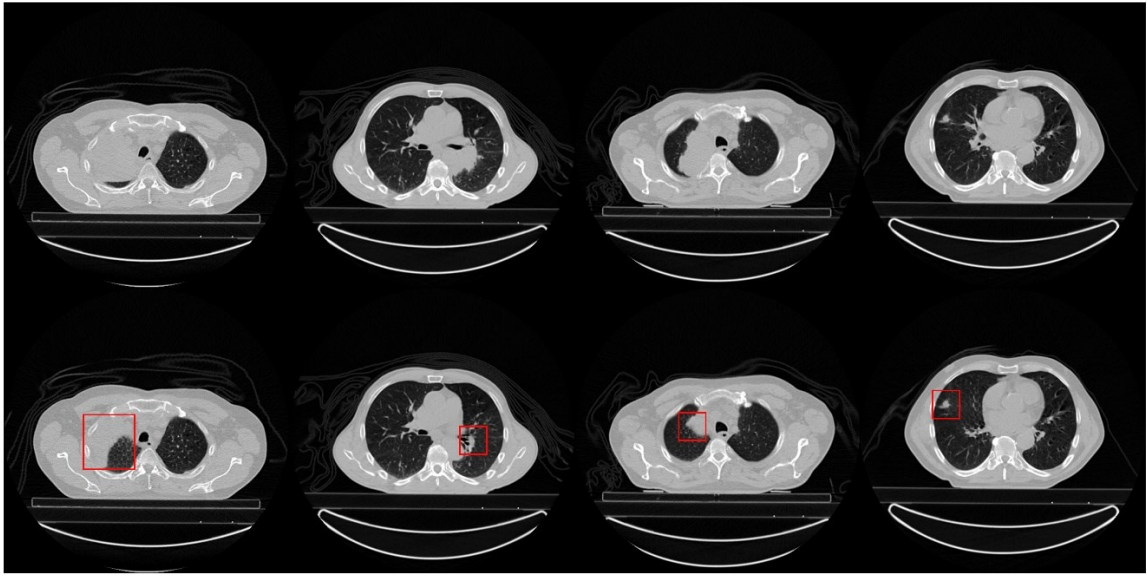

Figure 1.4. Examples of challenging LC tumors in CT images where the proposed autoinpaiting pipeline could partially remove the tumors (the first three examples from left to right), or it mistakenly removes the healthy structures (the last image). The first row shows the original tumoral slices, and the second row depicts the results of autoinpainting pipeline. The parts of the tumors which were not removed are highlighted in red bounding boxes.

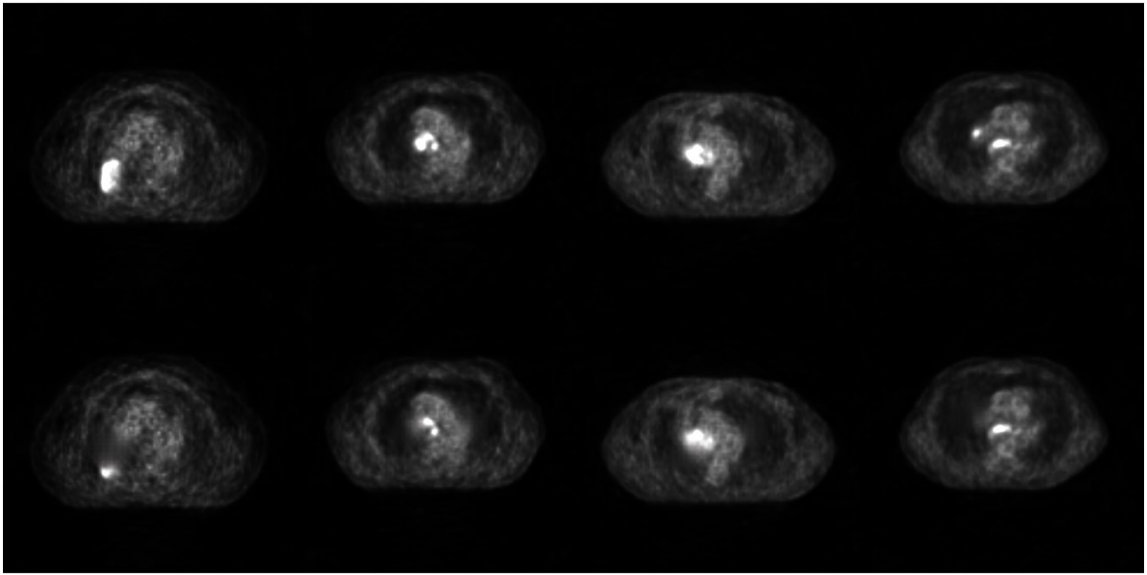

Figure 1.5. Examples of PET images where the tumors were partially removed by the proposed autoinpainting pipeline. The first row shows the original tumoral slices, and the second row depicts the results of autoinpainting pipeline.

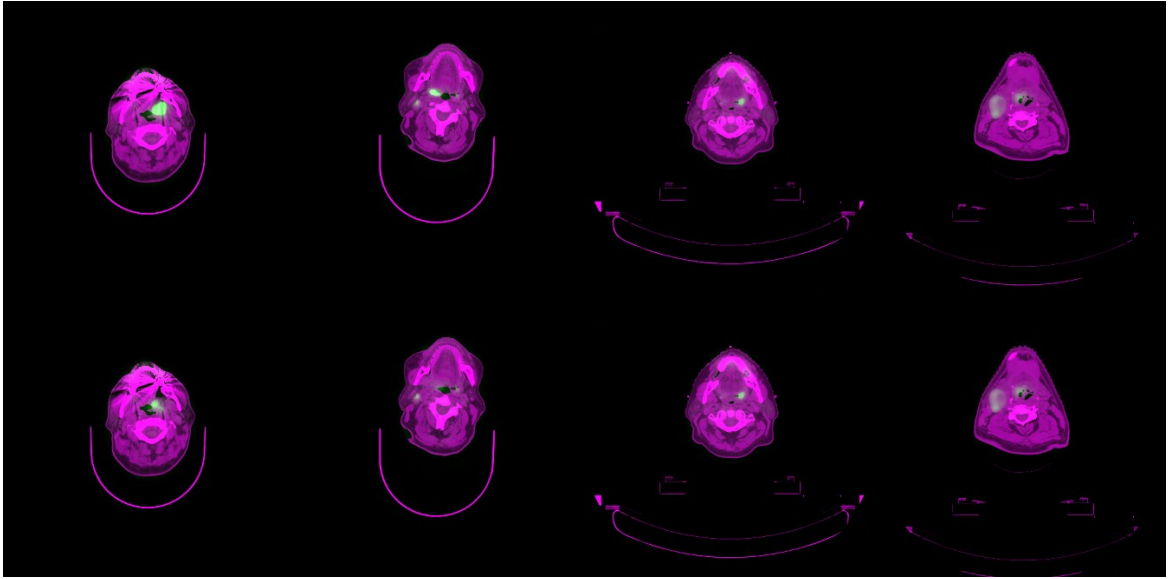

Figure 1.6. Examples of HN tumors in PET-CT images where the autoinpainting pipeline failed to completely substitute the tumoral regions with healthy anatomies. The first row shows the original tumoral slices, and the second row depicts the results of autoinpainting pipeline.

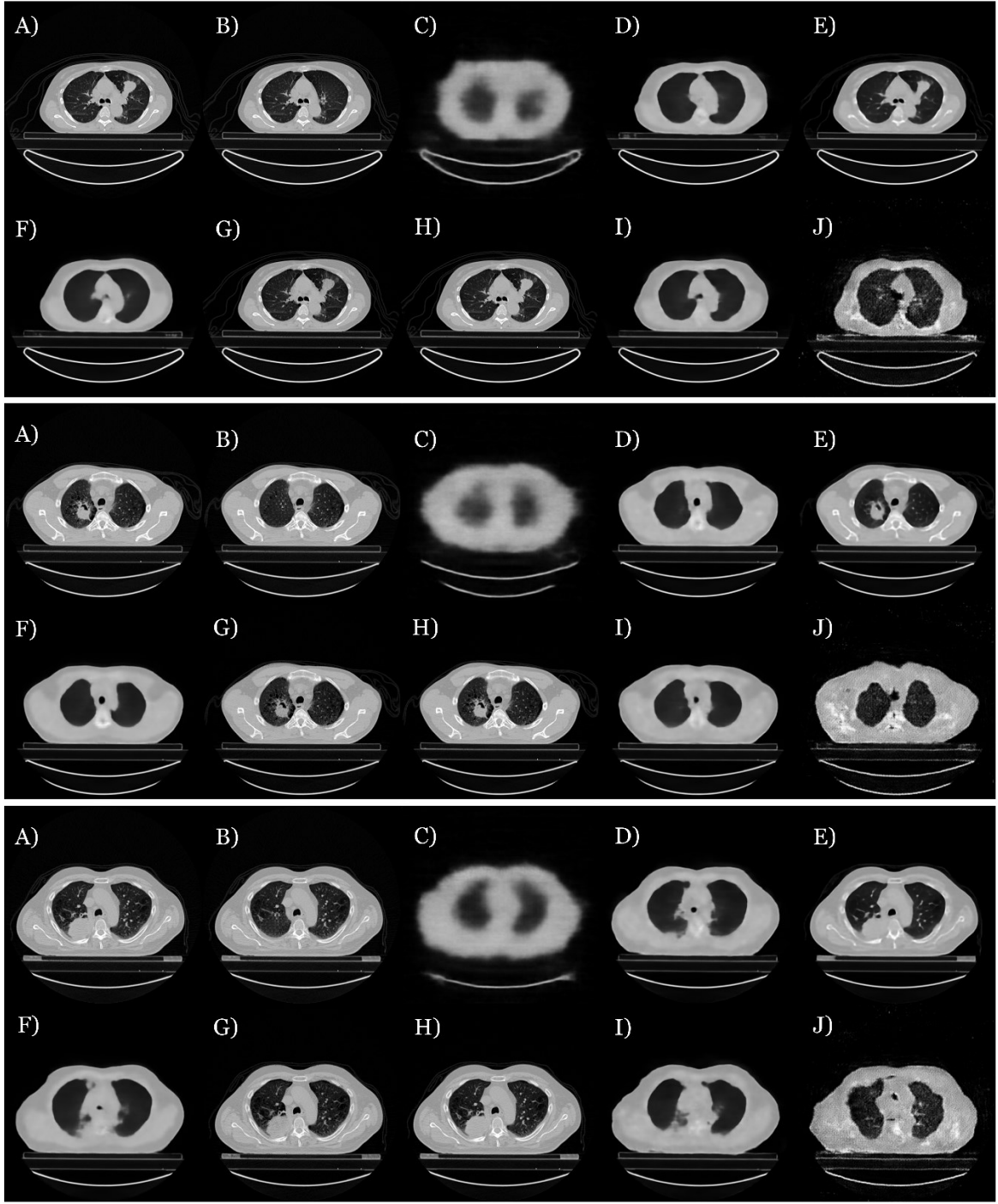

Figure 1.7. Qualitative comparison between the performance of the proposed autoinpainting pipeline and the other eight UAD models to learn the appearance of healthy anatomy of lungs in CT images. Each three sets of images include: A) original tumor slice, B) proposed autoinpainted image, C) adversarial autoencoder result, D) dense autoencoder result, E) spatial autoencoder result, F) variational autoencoder result, G) context-encoding variational autoencoder result, H) Gaussian mixture variational autoencoder result, I) context-encoding autoencoder result, and J) Fast-Anomaly GAN.

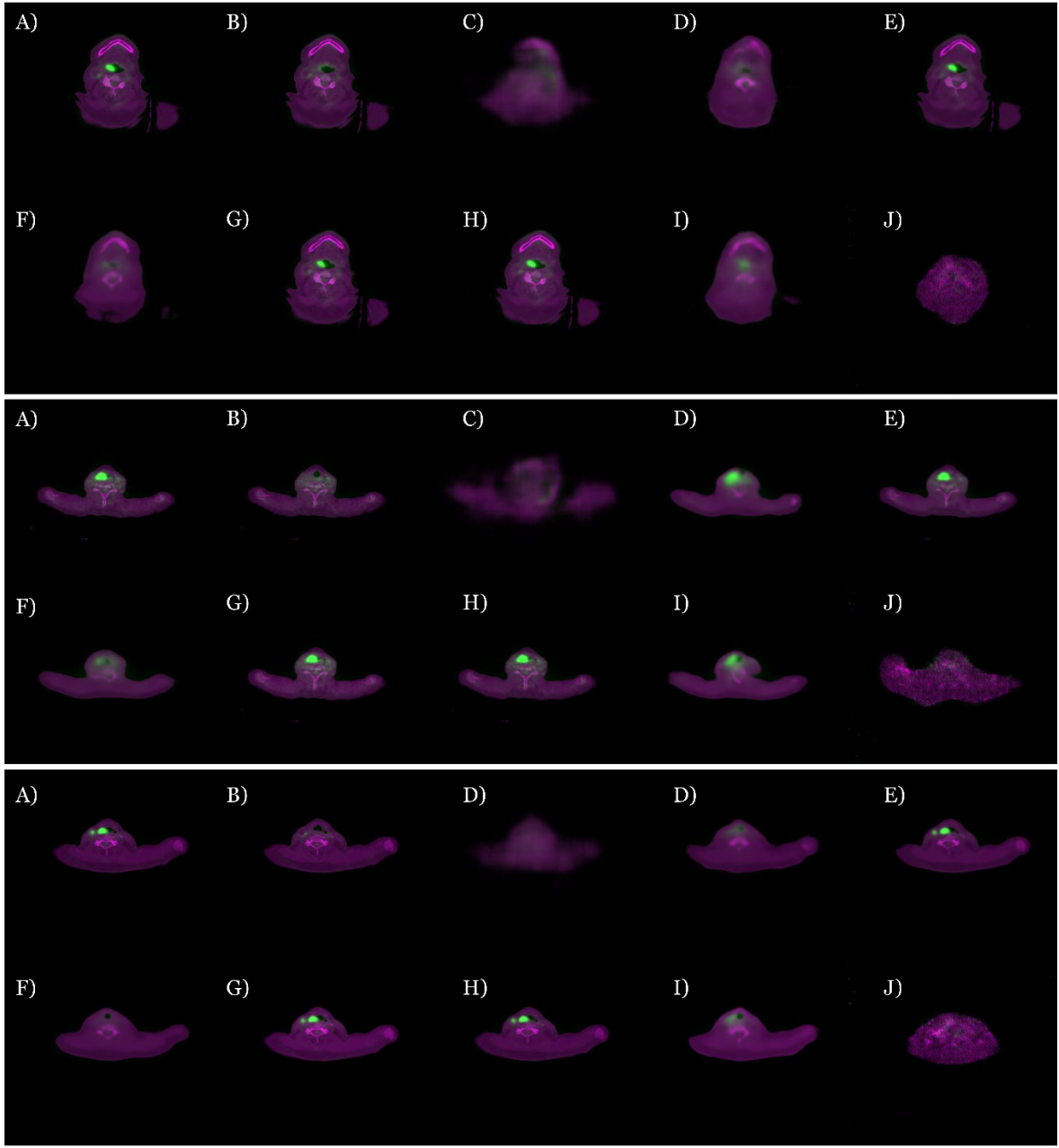

Figure 1.8. Qualitative comparison between the performance of the proposed autoinpainting pipeline and the other eight UAD models to learn the appearance of healthy anatomy of the head-and-neck region in PET-CT images. Each three sets of images include: A) original tumor slice, B) proposed autoinpainted image, C) adversarial autoencoder result, D) dense autoencoder result, E) spatial autoencoder result, F) variational autoencoder result, G) context-encoding variational autoencoder result, H) Gaussian mixture variational autoencoder result, I) context-encoding autoencoder result, and J) Fast-Anomaly GAN.

## 2) Ablation Study

### 2.1) Objective Function

The original objective function which was used for the Pconv model is:

$$\mathcal{L}_{total} = 1\mathcal{L}_{valid} + 6\mathcal{L}_{hole} + 0.05\mathcal{L}_{perceptual} + 120(\mathcal{L}_{style_{out}} + \mathcal{L}_{style_{comp}}) + 0.1\mathcal{L}_{tv}$$

In this study, the general framework of the objective function developed for the proposed Gconv$_{Lap}$ model is:

$$\mathcal{L}_{total} = C_1\mathcal{L}_{valid} + C_2\mathcal{L}_{hole} + C_3\mathcal{L}_{perceptual} + C_4(\mathcal{L}_{style_{out}} + \mathcal{L}_{style_{comp}}) + C_5\mathcal{L}_{tv} + C_6\mathcal{L}_{lap}$$

And the specific weighting coefficients of different terms are:

$$\mathcal{L}_{total} = 30\mathcal{L}_{valid} + 240\mathcal{L}_{hole} + 0.2\mathcal{L}_{perceptual} + 0.05(\mathcal{L}_{style_{out}} + \mathcal{L}_{style_{comp}}) + 250\mathcal{L}_{tv} + 20\mathcal{L}_{lap}$$

An extensive ablation study was conducted to determine the proper coefficients ($C_i$) of the above-mentioned loss function. In the following, a summary of the results is reported.

The coefficient of different terms of the objective function was, first, determined based on the idea that all the terms should contribute to the optimization equally so that none of them could outweigh the others. Accordingly, the following loss function was initiated:

$$\mathcal{L}_{total} = 20\mathcal{L}_{valid} + 120\mathcal{L}_{hole} + 0.1\mathcal{L}_{perceptual} + 0.015(\mathcal{L}_{style_{out}} + \mathcal{L}_{style_{comp}}) + 250\mathcal{L}_{tv} + 20\mathcal{L}_{lap}$$

Then, for each of the $C_i$ a set of values were set around the initial coefficients, and the model was trained 100 epochs for each setting independently. The image similarity metrics over a test set of 2000 images were quantified to determine the optimal values.

Table 2.1 shows the results of comparing the numerical metrics between the original Pconv model and the initial guess for the coefficients of the proposed model:

Table 2.1. The numerical comparison between the proposed model with the initial guess of loss term coefficients and the original Pconv model

| Experiment Name | Quantitative Metrics ($\mu\pm\sigma$) | | |
|---|---|---|---|
| | MSE | PSNR | SSIM |
| Original Pconv coef. | 117.460±59.841 | 28.004±2.277 | 0.884±0.021 |
| All coef. set to 1 | 78.775±47.747 | 30.046±2.925 | 0.930±0.022 |
| Initial guess without Laplasian loss | 64.009±44.861 | 31.317±3.598 | 0.943±0.021 |
| Initial guess with Laplassian loss | 61.922±44.460 | 31.471±3.586 | 0.946±0.020 |

Table 2.2 shows the effect of changing the coefficient $C_1$ i.e.:

$$\mathcal{L}_{total} = C_1\mathcal{L}_{valid} + 120\mathcal{L}_{hole} + 0.1\mathcal{L}_{perceptual} + 0.015(\mathcal{L}_{style_{out}} + 2\mathcal{L}_{style_{comp}}) + 250\mathcal{L}_{tv} + 20\mathcal{L}_{lap}$$

Table 2.2. The impact of changing the $C_1$ coefficients on the model performance. The candidate values for further experiments are marked in bold.

| $C_1$ variable | Quantitative Metrics ($\mu\pm\sigma$) | | |
|---|---|---|---|
| | MSE | PSNR | SSIM |
| **0** | 61.666±39.895 | 31.613±3.478 | 0.952±0.020 |
| 1 | 61.009±47.244 | 31.751±3.928 | 0.951±0.019 |
| 10 | 66.820±45.263 | 31.025±3.438 | 0.952±0.020 |
| 20 | 71.907±50.929 | 30.834±3.634 | 0.945±0.021 |
| **30** | 62.626±40.826 | 31.257±3.361 | 0.947±0.020 |
| 50 | 65.056±43.748 | 31.140±3.431 | 0.949±0.019 |

Table 2.3. shows the effect of changing the coefficient $C_2$ i.e.:

$$\mathcal{L}_{total} = 20\mathcal{L}_{valid} + C_2\mathcal{L}_{hole} + 0.1\mathcal{L}_{perceptual} + 0.015(\mathcal{L}_{style_{out}} + 2\mathcal{L}_{style_{comp}}) + 250\mathcal{L}_{tv} + 20\mathcal{L}_{lap}$$

Table 2.3. The impact of changing the $C_2$ coefficients on the model performance. The candidate values for further experiments are marked in bold.

| $C_2$ variable | Quantitative Metrics ($\mu\pm\sigma$) | | |
|---|---|---|---|
| | MSE | PSNR | SSIM |
| 0 | 61.368±43.172 | 31.430±3.449 | 0.951±0.019 |
| 1 | 63.964±44.135 | 31.246±3.473 | 0.950±0.021 |
| 10 | 61.220±45.115 | 31.573±3.605 | 0.949±0.020 |
| **60** | **59.938±40.708** | **31.434±3.290** | **0.948±0.020** |
| 120 | 72.804±46.925 | 30.600±3.446 | 0.946±0.020 |
| **240** | **58.955±41.414** | **31.624±3.484** | **0.950±0.020** |
| 500 | 62.696±40.600 | 31.229±3.545 | 0.949±0.019 |

Table 2.4. shows the effect of changing the coefficient $C_3$ i.e.:

$$\mathcal{L}_{total} = 20\mathcal{L}_{valid} + 120\mathcal{L}_{hole} + C_3\mathcal{L}_{perceptual} + 0.015(\mathcal{L}_{style_{out}} + 2\mathcal{L}_{style_{comp}}) + 250\mathcal{L}_{tv} + 20\mathcal{L}_{lap}$$

Table 2.4. The impact of changing the $C_3$ coefficients on the model performance. The candidate values for further experiments are marked in bold.

| $C_3$ variable | Quantitative Metrics ($\mu\pm\sigma$) | | |
|---|---|---|---|
| | MSE | PSNR | SSIM |
| 0 | 70.880±46.704 | 30.678±3.244 | 0.934±0.021 |
| 0.05 | 61.982±44.739 | 31.468±3.595 | 0.947±0.019 |
| 0.1 | 62.816±43.824 | 31.368±3.542 | 0.950±0.021 |
| **0.2** | **58.224±44.404** | **31.950±3.935** | **0.953±0.021** |

Table 2.5. shows the effect of changing the coefficient $C_4$ i.e.:

$$\mathcal{L}_{total} = 20\mathcal{L}_{valid} + 120\mathcal{L}_{hole} + 0.1\mathcal{L}_{perceptual} + C_4(\mathcal{L}_{style_{out}} + \mathcal{L}_{style_{comp}}) + 250\mathcal{L}_{tv} + 20\mathcal{L}_{lap}$$

Table 2.5. The impact of changing the $C_4$ coefficients on the model performance. The candidate values for further experiments are marked in bold.

| $C_4$ variable | Quantitative Metrics ($\mu\pm\sigma$) | | |
|---|---|---|---|
| | MSE | PSNR | SSIM |
| **0** | **66.653±39.837** | **31.003±3.487** | **0.945±0.019** |
| **0.05** | **67.448±45.412** | **30.956±3.384** | **0.949±0.020** |
| **0.1** | **66.550±44.078** | **31.056±3.481** | **0.943±0.021** |
| 0.2 | 80.045±53.061 | 30.166±3.272 | 0.942±0.021 |
| 1 | 83.138±53.245 | 29.926±3.127 | 0.928±0.022 |
| 10 | 143.102±62.211 | 26.998±1.969 | 0.874±0.021 |
| 60 | 138.358±67.401 | 27.279±2.292 | 0.894±0.021 |

Table 2.6 shows the effect of changing the coefficient $C_5$ i.e.:`

$$\mathcal{L}_{total} = 20\mathcal{L}_{valid} + 120\mathcal{L}_{hole} + 0.1\mathcal{L}_{perceptual} + 0.015(\mathcal{L}_{style_{out}} + \mathcal{L}_{style_{comp}}) + C_5\mathcal{L}_{tv} + 20\mathcal{L}_{lap}$$

Table 2.6. The impact of changing the $C_5$ coefficients on the model performance. The candidate values for further experiments are marked in bold.

| $C_5$ variable | Quantitative Metrics ($\mu \pm \sigma$) | | |
|---|---|---|---|
| | MSE | PSNR | SSIM |
| 0 | 65.480±38.833 | 31.203±3.218 | 0.941±0.018 |
| 0.1 | 61.467±41.109 | 31.374±3.420 | 0.950±0.020 |
| 1 | 64.616±43.502 | 31.052±3.167 | 0.945±0.019 |
| 10 | 65.461±50.113 | 31.419±3.835 | 0.949±0.021 |
| **100** | **59.500±41.086** | **31.639±3.738** | **0.949±0.021** |
| **250** | **58.340±41.086** | **31.769±3.738** | **0.949±0.019** |
| 500 | 65.933±48.985 | 31.373±3.926 | 0.948±0.020 |

Finally, Table 2.7. shows the effect of changing the coefficient $C_6$ i.e.:

$$\mathcal{L}_{total} = 20\mathcal{L}_{valid} + 120\mathcal{L}_{hole} + 0.1\mathcal{L}_{perceptual} + 0.015(\mathcal{L}_{style_{out}} + \mathcal{L}_{style_{comp}}) + 250\mathcal{L}_{tv} + C_6\mathcal{L}_{lap}$$

Table 2.7. The impact of changing the $C_6$ coefficients on the model performance. The candidate values for further experiments are marked in bold.

| $C_6$ variable | Quantitative Metrics ($\mu \pm \sigma$) | | |
|---|---|---|---|
| | MSE | PSNR | SSIM |
| 0 | 65.924±42.228 | 31.213±3.533 | 0.940±0.020 |
| **1** | **63.426±44.382** | **31.289±3.453** | **0.946±0.019** |
| 10 | 68.367±41.182 | 30.699±3.061 | 0.943±0.018 |
| **20** | **63.121±45.666** | **31.394±3.585** | **0.950±0.020** |
| 50 | 66.587±44.098 | 30.945±3.246 | 0.947±0.019 |
| 100 | 64.476±40.940 | 31.268±3.407 | 0.948±0.020 |

From the described conducted experiments, the final weight candidates for each of the loss terms will be:

$$\mathcal{L}_{total} = \begin{Bmatrix} 0 \\ 1 \\ 30 \end{Bmatrix} \mathcal{L}_{valid} + \begin{Bmatrix} 60 \\ 240 \end{Bmatrix} \mathcal{L}_{hole} + 0.2\mathcal{L}_{perceptual} + \begin{Bmatrix} 0 \\ 0.05 \\ 0.1 \end{Bmatrix} (\mathcal{L}_{style_{out}} + \mathcal{L}_{style_{comp}}) + \begin{Bmatrix} 100 \\ 250 \end{Bmatrix} \mathcal{L}_{tv} + \begin{Bmatrix} 1 \\ 20 \end{Bmatrix} \mathcal{L}_{lap}$$

Therefore, to determine the optimal values of the weight coefficients, a set of independent experiments was examined by setting the different combinations of the weight candidates (Table 2.8).

Table 2.8. The impact of different combinations of weighting coefficients on the model performance. The final candidate values are marked in bold.

| Experiment | Quantitative Metrics ($\mu \pm \sigma$) | | |
|---|---|---|---|
| | MSE | PSNR | SSIM |
| $0L_{valid}+60L_{hole}+0.2L_{per.}+0(L_{style_{out}}+L_{style_{comp}})+100L_{tv}+1L_{lap}$ | 56.235±36.519 | 31.749±3.614 | 0.954±0.027 |
| $1L_{valid}+60L_{hole}+0.2L_{per.}+0(L_{style_{out}}+L_{style_{comp}})+100L_{tv}+1L_{lap}$ | 53.175±29.341 | 31.837±4.267 | 0.957±0.031 |
| $1L_{valid}+60L_{hole}+0.2L_{per.}+0.1(L_{style_{out}}+L_{style_{comp}})+100L_{tv}+1L_{lap}$ | 48.591±38.947 | 32.830±4.101 | 0.960±0.019 |
| $1L_{valid}+60L_{hole}+0.2L_{per.}+0.1(L_{style_{out}}+L_{style_{comp}})+100L_{tv}+20L_{lap}$ | 48.413±38.700 | 32.959±4.339 | 0.961±0.020 |
| $1L_{valid}+240L_{hole}+0.2L_{per.}+0.1(L_{style_{out}}+L_{style_{comp}})+100L_{tv}+20L_{lap}$ | 52.182±29.381 | 31.862±3.962 | 0.961±0.041 |
| $30L_{valid}+60L_{hole}+0.2L_{per.}+0.1(L_{style_{out}}+L_{style_{comp}})+100L_{tv}+20L_{lap}$ | 50.439±36.468 | 32.463±3.830 | 0.962±0.019 |
| $30L_{valid}+240L_{hole}+0.2L_{per.}+0.1(L_{style_{out}}+L_{style_{comp}})+100L_{tv}+20L_{lap}$ | 47.938±36.107 | 32.768±3.960 | 0.963±0.018 |
| $30L_{valid}+60L_{hole}+0.2L_{per.}+0.05(L_{style_{out}}+2L_{style_{comp}})+250L_{tv}+20L_{lap}$ | 45.245±35.381 | 33.272±4.469 | 0.966±0.018 |
| **$30L_{valid}+240L_{hole}+0.2L_{per.}+0.05(L_{style_{out}}+L_{style_{comp}})+250L_{tv}+20L_{lap}$** | **44.290±33.785** | **33.271±4.267** | **0.966±0.018** |
| $30L_{valid}+240L_{hole}+0.2L_{per.}+0.05(L_{style_{out}}+L_{style_{comp}})+100L_{tv}+20L_{lap}$ | 46.238±34.697 | 33.013±4.154 | 0.963±0.020 |

Therefore, the set of weight coefficients led to inpainting the images with the highest quality was determined, and the final objective function is defined as:

$$\mathcal{L}_{total} = 30\mathcal{L}_{valid} + 240\mathcal{L}_{hole} + 0.2\mathcal{L}_{perceptual} + 0.05(\mathcal{L}_{style_{out}} + \mathcal{L}_{style_{comp}}) + 250\mathcal{L}_{tv} + 20\mathcal{L}_{lap}$$

2.2) Autoinpainting

The protocols of the proposed autoinpainting pipeline require the choice of two hyperparameters: 1) the radius of the moving windows (circles) and 2) the number of top candidate regions. Accordingly, a set of independent experiments were examined to determine the optimal values of these two parameters.

To specify the radius of the moving circles and the number of top candidates, a range of different values was studied, and the effect of these values on the segmentation accuracy was quantified (Table 2.9.).

Table 2.9. The impact of changing the radius of the moving circles and the number of top candidate regions on the segmentation accuracy. The final candidate values are marked in bold.

| Top Candidates - Circle Radius | Segmentation Metrics ($\mu \pm \sigma$) | | |
|---|---|---|---|
| | Dice | Sensitivity | Specificity |
| 1-23 | 0.305±0.231 | 0.291±0.153 | 0.982±0.001 |
| 1-25 | 0.308±0.189 | 0.289±0.114 | 0.984±0.001 |
| 1-27 | 0.312±0.233 | 0.301±0.149 | 0.984±0.001 |
| 1-29 | 0.321±0.194 | 0.309±0.213 | 0.984±0.001 |
| 1-31 | 0.328±0.183 | 0.305±0.142 | 0.986±0.001 |
| 2-23 | 0.395±0.192 | 0.381±0.183 | 0.999±0.000 |
| 2-25 | 0.393±0.159 | 0.378±0.128 | 0.999±0.000 |
| 2-27 | 0.410±0.193 | 0.392±0.176 | 0.999±0.000 |
| 2-29 | 0.401±0.203 | 0.389±0.127 | 0.999±0.000 |
| 2-31 | 0.403±0.143 | 0.395±0.142 | 0.999±0.000 |
| 3-23 | 0.422+0.167 | 0.416+0.160 | 0.999+0.000 |
| 3-25 | 0.429+0.170 | 0.412+0.167 | 0.999+0.000 |
| **3-27** | **0.437+0.172** | **0.419+0.171** | **0.999+0.000** |
| 3-29 | 0.433+0.174 | 0.415+0.174 | 0.999+0.000 |
| 3-31 | 0.419+0.174 | 0.412+0.176 | 0.999+0.000 |

Therefore, the radius of the moving circles was set as 27 pixels and the number of top candidate regions was determined as 3.